\documentclass{article}

\usepackage[preprint]{neurips_2023}

\usepackage[utf8]{inputenc} % allow utf-8 input
\usepackage[T1]{fontenc}    % use 8-bit T1 fonts
\usepackage[colorlinks]{hyperref}

\usepackage{url}            % simple URL typesetting
\usepackage{booktabs}       % professional-quality tables
\usepackage{amsfonts}       % blackboard math symbols
\usepackage{nicefrac}       % compact symbols for 1/2, etc.
\usepackage{microtype}      % microtypography

\usepackage{multirow}
\usepackage{threeparttable}
\usepackage{graphicx}
\usepackage{booktabs} % professional-quality tables
\usepackage{dsfont}
\usepackage{nicefrac}
\usepackage{pifont}% http://ctan.org/pkg/pifont  % for cmark
\usepackage{flushend}
\usepackage{caption}
\usepackage{wrapfig}
\usepackage{array}
\usepackage{siunitx}
\usepackage[export]{adjustbox}

\usepackage{graphicx}
\usepackage{booktabs}
\usepackage{threeparttable}
\usepackage[accsupp]{axessibility}  % Improves PDF readability for those with disabilities.
\usepackage{tikz}
\usepackage{comment}
\usepackage{amsmath,amssymb} % define this before the line numbering.
\usepackage{xcolor}

\makeatletter
\newcommand{\figcaption}[1]{\def\@captype{figure}\caption{#1}}
\newcommand{\tblcaption}[1]{\def\@captype{table}\caption{#1}}
\makeatother
\usepackage{orcidlink}

\newcommand\blfootnote[1]{%
\begingroup
\renewcommand\thefootnote{}\footnote{#1}%
\addtocounter{footnote}{-1}%
\endgroup
}

\title{MMVR: Millimeter-wave Multi-View Radar Dataset and Benchmark for Indoor Perception}

\author{
M. Mahbubur Rahman\orcidlink{0000-0002-4883-9767}$^{\ast, \dag}$,
Ryoma Yataka\orcidlink{0009-0004-7311-6431}$^{\ast, \sharp}$,
Sorachi Kato\orcidlink{0000-0002-6896-5293}$^{\ast, \dag}$,
Pu (Perry) Wang\orcidlink{0000-0002-4718-3102}$^{\ast, \ddag}$, 
\\
\textbf{Peizhao Li}\orcidlink{0000-0001-5590-3977}$^{\dag}$,
\textbf{Adriano Cardace}\orcidlink{0000-0002-0584-3109}$^{\dag}$,
\textbf{Petros Boufounos}\orcidlink{0000-0003-1369-0947}
  \\ \\
Mitsubishi Electric Research Laboratories (MERL) \\
Cambridge, MA 02319, USA
}
\newcommand{\ba}{\begin{array}}
\newcommand{\ea}{\end{array}}
\newcommand{\be}{\begin{displaymath}}
\newcommand{\ee}{\end{displaymath}}
\newcommand{\ben}{\begin{equation}}
\newcommand{\een}{\end{equation}}
\newcommand{\bena}{\begin{eqnarray}}
\newcommand{\eena}{\end{eqnarray}}
\newcommand{\beqa}{\begin{eqnarray*}}
\newcommand{\enqa}{\end{eqnarray*}}

\newcommand{\bc}{\begin{center}}
\newcommand{\ec}{\end{center}}
\newcommand{\bi}{\begin{itemize}}
\newcommand{\ei}{\end{itemize}}
\newcommand{\benu}{\begin{enumerate}}
\newcommand{\eenu}{\end{enumerate}}
\newcommand{\bdes}{\begin{description}}
\newcommand{\edes}{\end{description}}
\newcommand{\bt}{\begin{tabular}}
\newcommand{\et}{\end{tabular}}

\newcommand \mubf{\mbox{\boldmath$\mu$\unboldmath}}

\newcommand \tbf{{\bf t}}

\newcommand \Hbf{{\bf H}}

\newcommand \Rbf{{\bf R}}
\newcommand \Sbf{{\bf S}}

\newcommand \Ubf{{\bf U}}
\newcommand \Vbf{{\bf V}}

\newcommand{\circlambda}{\mbox{$\Lambda$
             \kern-.85em\raise1.5ex
             \hbox{$\scriptstyle{\circ}$}}\,}

\renewcommand \mubf{\boldsymbol{\mu}}

\begin{document}
\maketitle

%%%%%%%%%%%%%%%%%%%%%%%%%%%%%%%%%%%%%%%%%%%
\blfootnote{$^\ast$: Equal contribution. $^\dag$: The work of M. Rahman (Univ. of Alabama, USA), S. Kato (Osaka Univ., Japan), P. Li (Brandeis Univ., USA), and A. Cardace (Univ. of Bologna, Italy) was done during their internship at MERL. $^\sharp$: The work was done as a visiting scientist from Mitsubishi Electric Corporation, Japan. $^\ddag$: Project Lead.}

%%%%%%%%%%%%%%%%%%%%%%%%%%%%%%%%%%%%%%%%%%%%%%%%%%%%%%%%%%%%%%%%%%%%%
\begin{abstract}
Compared with an extensive list of automotive radar datasets that support autonomous driving, indoor radar datasets are scarce at a smaller scale in the format of low-resolution radar point clouds and usually under an open-space single-room setting. In this paper, we scale up indoor radar data collection using multi-view high-resolution radar heatmap in a multi-day, multi-room, and multi-subject setting, with an emphasis on the diversity of environment and subjects. Referred to as the millimeter-wave multi-view radar (\textbf{MMVR}) dataset, it consists of $345$K multi-view radar frames collected from $25$ human subjects over $6$ different rooms, $446$K annotated bounding boxes/segmentation instances, and $7.59$ million annotated keypoints to support three major perception tasks of object detection, pose estimation, and instance segmentation, respectively.  For each task, we report performance benchmarks under two protocols: a single subject in an open space and multiple subjects in several cluttered rooms with two data splits: random split and cross-environment split over $395$ 1-min data segments. We anticipate that MMVR facilitates indoor radar perception development for indoor vehicle (robot/humanoid) navigation, building energy management, and elderly care for better efficiency, user experience, and safety. The MMVR dataset is available at \url{https://doi.org/10.5281/zenodo.12611978}.
\end{abstract}

%%%%%%%%%%%%%%%%%%%%%%%%%%%%%%%%%%%%%%%%%%%%%%%%%%%%%%%%%%%%%%%%%%%%%%%%%%%%%%%%%%%%%%%%%%%%%%%%%%%%%%%%%%%%%%%%
\section{Introduction}
\label{sec:intro}
Compared with popular automotive radar perception efforts and datasets~\cite{nuscenes20, RadarScenes21, Radiate21, Oxford20, Mulran20, CARRADA21, CRUW21, RaDICaL21, RADIal21, VoD22, Kradar22, tempoRadar22, BilikLongman19, PandharipandeCheng23, SIRA24}, indoor radar perception receives less attention but it is essential in applications such as indoor vehicle (robot/humanoid) navigation, building energy management, and elderly care under low light and emergence (smoke, fire, and dust) situations with benefits of low device cost and less privacy concerns.  As a result, its development lags and the open datasets are limited in terms of size, annotation, tasks, and benchmarks. 

\begin{figure}[t]
  \centering
        \includegraphics[width=0.26\textwidth]{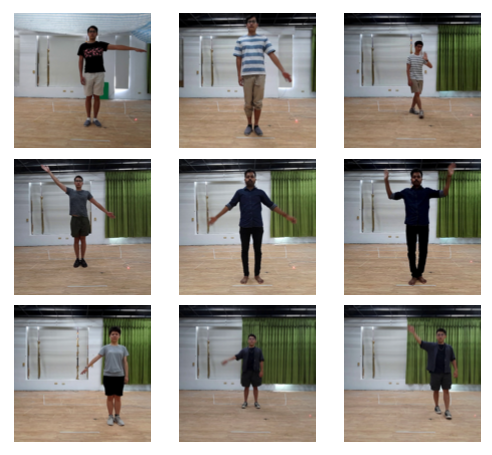} 
\hfill 
        \includegraphics[width=0.32\textwidth]{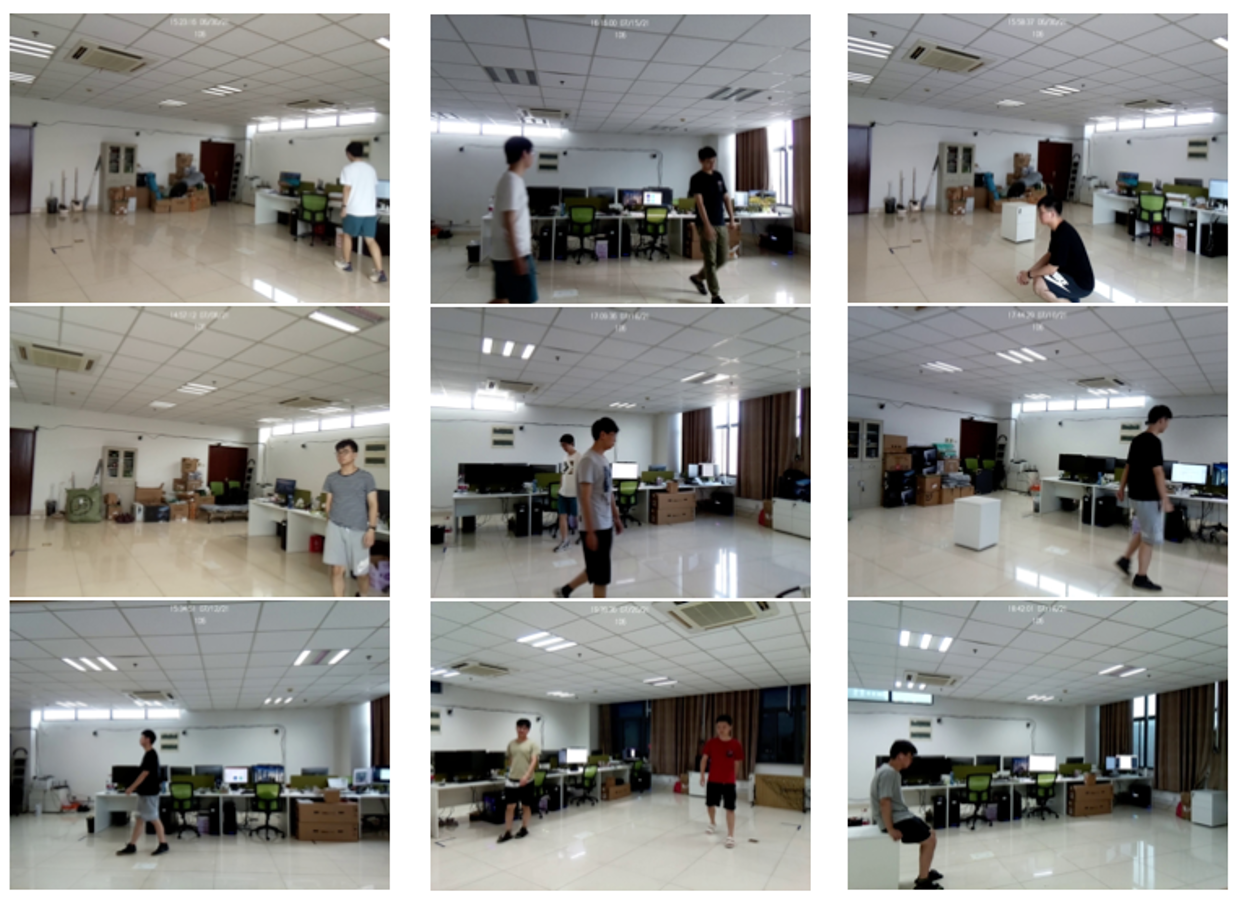}
\hfill 
        \includegraphics[width=0.31\textwidth]{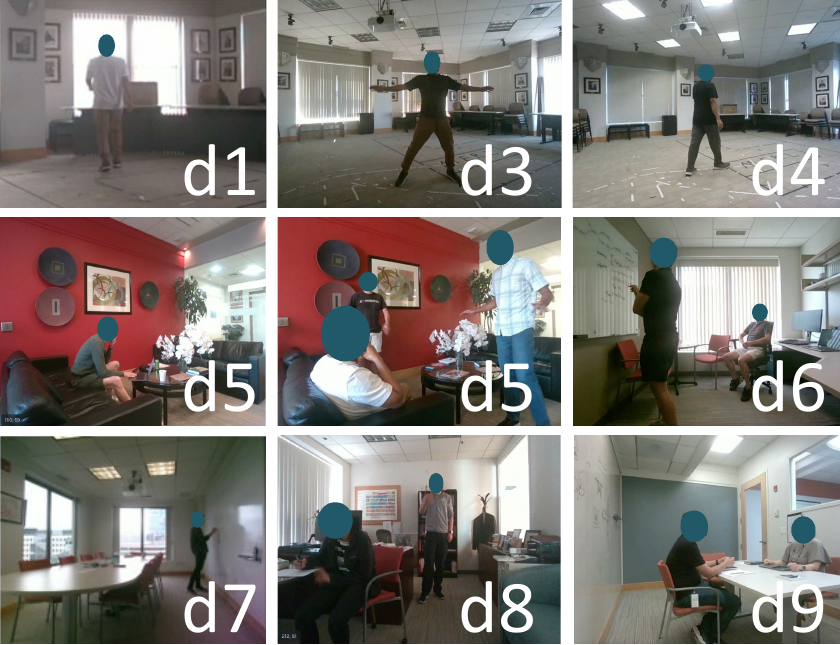} \\
          (a) HuPR (open space)  \quad\quad\quad (b) HIBER (open foreground) \quad\quad\quad (c) MMVR (cluttered space)
  \caption{Snapshots of indoor radar \textbf{heatmap} datasets in Table~\ref{table: heatmap}. HuPR~(\cite{HuPR23}) and HIBER~(\cite{RFMask23}) were collected at a single location and, respectively, multiple locations inside a single open-space/open-foreground room. Our MMVR was collected in multiple locations over multiple open-foreground (d1-d4) and cluttered (d5-d9) rooms.  } 
  \label{fig:snapshots}
\end{figure}

A few lines of effort have explored radar signals and processed results for indoor perception. The earliest work is RF-Pose~(\cite{RFPose18}) using a T-shape antenna array~(\cite{RFCapture15}) ($16$ horizontal and $4$ vertical antennas) in the $5-7$ \si{\giga\hertz} band. It takes two orthogonal radar heatmaps generated by the T-shape array as the input, uses a convolution-based autoencoder network to fuse features from the two radar views, and regresses multi-person keypoints for 2D image-plane pose estimation. It is noted that RF-Pose is not publicly accessible. More recently, HuPR has made a similar effort collecting two orthogonal LR radar heatmaps for action classification in an empty room (see Fig.~\ref{fig:snapshots}~(a)) and pose estimation and publicly releases the dataset~(\cite{HuPR23}). 
Due to the fact that their angular (azimuth and elevation) resolution is limited to $15^\circ$~(\cite{RFPose18, HuPR23}), we refer to RF-Pose and HuPR as \textbf{multi-view low-resolution (LR) radar heatmap} datasets in Table~\ref{table:datasets}. 

\begin{figure}[t]
    \centering
    \includegraphics[width=0.99\textwidth]{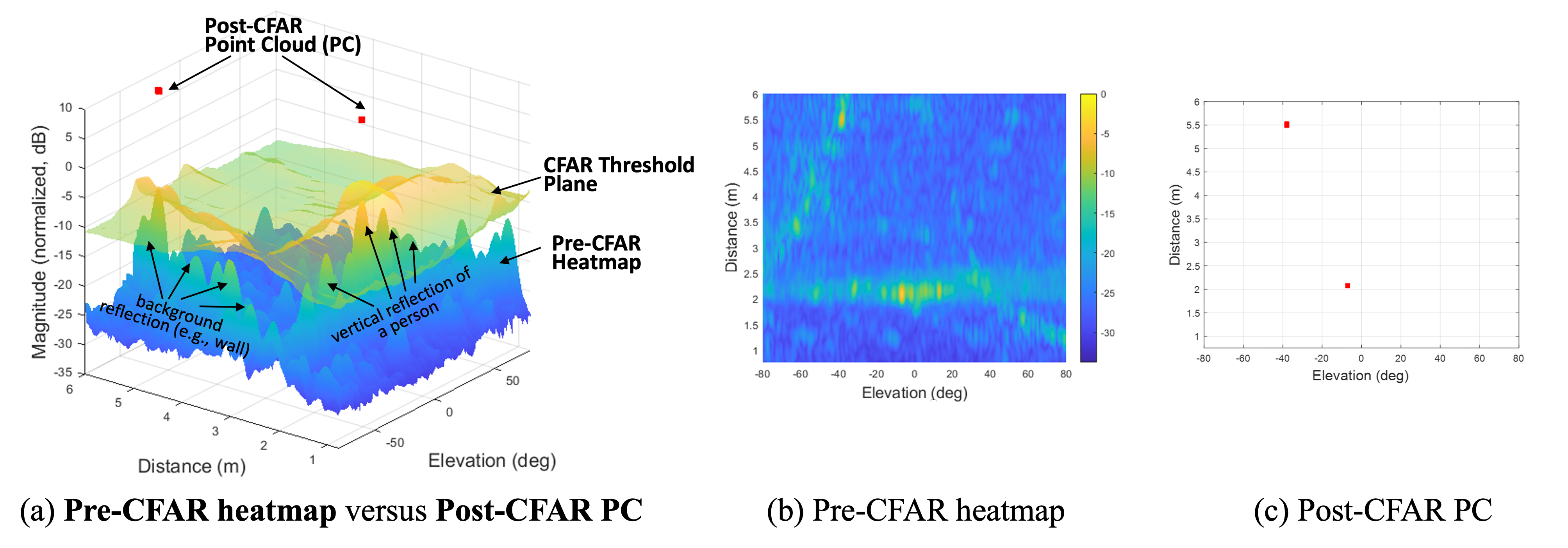} 
    \caption{Radar signal representations: pre-CFAR heatmap versus post-CFAR point cloud (PC) corresponding to a scene in our MMVR dataset. The heatmap ((a) for a 3D view and (b) for the top-down view)  shows an extended vertical profile of a subject in terms of multiple clustered reflections over the elevation (height) domain. In contrast, the CFAR operation compares the heatmap with a threshold plane (the semi-transparent surface in (a)) to declare a few detection points (red squares in (a) and (c)), greatly suppressing weaker reflections from the subject. } 
    \label{fig:cfar}
\end{figure}

\begin{table}[b]
    \centering
    \caption{\textbf{Indoor Radar Perception} Datasets.}
    \resizebox{1.0\columnwidth}{!}{%
    \begin{threeparttable}
    \begin{tabular}{ccccccccc}
    \toprule
    \multirow{1}*{Datasets} & \multirow{1}*{Year} & \multirow{1}*{Sensor}  & \multirow{1}*{Views} & \multirow{1}*{Data} & \multirow{1}*{Tasks}  & \multirow{1}*{Size} & \multirow{1}*{Public}\\
    \midrule
    RadHAR~(\cite{RadHAR19}) & 2019 & LR (low-resolution) & Single & PC (Point Cloud) & Action & $167$K & $\checkmark$\\
    mm-Pose~(\cite{mmPose20}) & 2020 & LR & Multi & PC & Action, Pose  & $40$K & \ding{53}\\
    mmMesh~(\cite{mmMesh21}) & 2021 & LR & Single & PC & 3D Mesh & $480$K & \ding{53}\\
    mRI~(\cite{mRI22}) & 2022 & LR & Single & PC & Action, Pose  & $160$K & $\checkmark$ \\
    MM-Fi~(\cite{mmFi23}) & 2023 & LR & Single & PC & Action, Pose  & $320$K & $\checkmark$\\
    \midrule
    RF-Pose~(\cite{RFPose18}) & 2018 & LR & Multi & Heatmap & Pose  & - & \ding{53}\\
    HuPR~(\cite{HuPR23}) & 2023 & LR & Multi & Heatmap & Pose & $141$K & $\checkmark$\\
    \midrule
    HIBER~(\cite{RFMask23}) & 2023 & HR (high-resolution) & Multi & Heatmap & Box, Pose, Seg. & $179$K & Partial\\
    \textbf{MMVR}(ours)  & 2024 & HR & Multi & Heatmap & Box, Pose, Seg.  & 345K & $\checkmark$\\
    \bottomrule
    \end{tabular}%
    \begin{tablenotes}
    \item[]LR: low-resolution radar with an angular resolution of $15^\circ$. 
    \item[]HR: high-resolution radar with an angular resolution of $1.3^\circ$. 
    \end{tablenotes}
    \end{threeparttable}
    }
    \label{table:datasets}
\end{table}

Following RF-Pose but with commercial radar sensors, other earlier efforts use TI's single-chip millimeter-wave (mmWave) radar consisting of $3$ transmitters and $4$ receivers in the $60-64$ \si{\giga\hertz} and $76-81$ \si{\giga\hertz} bands.  
These single-chip radar sensors achieve a comparable angular resolution of around $15^\circ$ as the RF-Pose due to a lower wavelength. Other than the multi-view radar heatmaps, indoor radar datasets such as RadHAR~(\cite{RadHAR19}), mm-Pose~(\cite{mmPose20}), mRI~(\cite{mRI22}), and more recently, MM-Fi~(\cite{mmFi23}) collect \textbf{single-view LR point cloud} (PC), usually along the horizontal orientation, as shown in the first block of Table~\ref{table:datasets}. 

The difference between PC and heatmap may have significant consequences on the downstream tasks such as the image-plane pose estimation and segmentation. This is illustrated in Fig.~\ref{fig:cfar}, where radar PC is obtained by applying the constant false alarm rate (CFAR) operation ~(\cite{Scharf1991_CFAR}) to the radar heatmap for a scene with a person moving at a distance of \SI{2}{\meter} in our MMVR dataset. The heatmap ((a) for a 3D view and (b) for the top-down view)  shows an extended vertical profile of a subject in terms of multiple clustered reflections over the elevation (height) domain. In contrast, simple CFAR operations compare the heatmap with an adaptively determined threshold plane (the semi-transparent surface in (a)) to declare a few detection points (red squares in (a) and (c)), greatly suppressing weaker reflections from the subject and missing fine-grained features for challenging tasks. As shown in Table~\ref{table:datasets}, these single-view LR point cloud datasets are mainly used for tasks such as action classification~(\cite{RadHAR19, mmPose20, mRI22, mmFi23}). 

To enable more fine-grained radar feature extraction, \textbf{multi-view high-resolution (HR) heatmaps} have been considered more recently for indoor radar perception such as HIBER~(\cite{RFMask23}). However, HIBER was collected at multiple locations within the {same room with open foreground} (see Fig.~\ref{fig:snapshots}~(b)). And there is no natural occlusion (table, chair, furniture) in front of subjects. To advance indoor radar perception in more complex environments (multiple rooms with cluttered space) and motivate challenging indoor downstream tasks using radar signals, we introduce a large-scale indoor radar perception dataset: Millimeter-wave Multi-View Radar (\textbf{MMVR}) with the following main contributions:
\begin{itemize}
    \item First, our dataset MMVR has $\mathbf{345}$\textbf{K} data frames collected from $\mathbf{25}$ human subjects over $\mathbf{6}$ different rooms (e.g, open/cluttered offices and meeting rooms) spanning over $\mathbf{9}$ separate days. 
    To the best of our knowledge, this is the largest open-source indoor radar dataset in a truly \textbf{multi-day}, \textbf{multi-room}, and \textbf{multi-subject} setting.
     \item Second, MMVR consists of $2$ parts: 1) $107.9$K data frames of \textbf{P1: Open Foreground} in a single open-foreground space with a single subject; see the first row of Fig.~\ref{fig:snapshots}~(c) for snapshots; and 2) $237.9$K data frames of \textbf{P2: Cluttered Space} in $5$ cluttered rooms with multiple subjects; see the second and third rows of Fig.~\ref{fig:snapshots}~(c). \textbf{P1} is used to establish the best possible radar perception benchmarks, while \textbf{P2} is designed for more challenging scenarios and for cross-environment and cross-subject generalization. 
    \item Third, MMVR has annotated about $\mathbf{446}$\textbf{K} bounding boxes, $\mathbf{7.59}$ \textbf{million} keypoints, and $\mathbf{446}$\textbf{K} segmentation instances (see Table~\ref{table: heatmap} for detailed dataset statistics). We leverage state-of-art RGB-based pipelines, i.e., Mask2Former~(\cite{mask2former}) and HRNets~(\cite{hrnet}), to generate high-confidence image-plane annotation labels (bounding boxes, keypoints, and segmentation pixels) with human curation involved to support three perception tasks:  1) object detection, 2) pose estimation, and 3) instance segmentation, respectively. 
    \item Finally, we re-implement or modify state-of-art radar perception baseline methods: RF-Pose~(\cite{RFPose18}) and RFMask~(\cite{RFMask23}), for all three tasks and establish benchmarks under both \textbf{P1} and \textbf{P2}.  Ablation studies show the impact of factors such as the number of input frames,  the number of radar views, and different data splits on all considered perception tasks. 
\end{itemize}

\begin{table}[b]
\centering
\caption{Comparison of multi-view heatmap-based indoor radar datasets.}
\resizebox{1.0\columnwidth}{!}{%
\begin{threeparttable}
\begin{tabular}{l|ccc|ccc|cccccc}
\toprule
 \multirow{2}{*}{\textbf{Dataset}} & \multicolumn{3}{c|}{\textbf{Resolution}} & \multicolumn{3}{c|}{\textbf{Annotations}} & \multicolumn{6}{c}{\textbf{Statistics}} \\
\cline{2-13}
                        & {Range} & {Azi.} & {Ele.} & {BBox} & {KP} & {Seg} & {Rooms}& {Subjects} & {Actions} & {Sequences} & {Frames} & {Room Setting} \\
\midrule
RF-Pose$^{\dag}$      & \SI{10.0}{\centi\meter} & $15^\circ$ & $15^\circ$  & \ding{53} & $\checkmark$ & \ding{53} & 50 & 100 & free-form  & - & - & - \\
HuPR      & \SI{4.8}{\centi\meter} & $15^\circ$ & $15^\circ$ & $141$K & $1.97$M & \ding{53} & 1 & 6 & static, walking & 235 & 141K & open space  \\
HIBER        & \SI{12.2}{\centi\meter} & $1.3^\circ$ & $1.3^\circ$ & ${231}${K}  & ${3.23}${M} & ${231}${K}  & 1 & 10 & free-form & 152 & 179K  & open foreground \\
\textbf{MMVR}(ours) & \SI{11.5}{\centi\meter} &  $1.3^\circ$ & $1.3^\circ$  & $\mathbf{446}$\textbf{K} & $\mathbf{7.59}$\textbf{M} & $\mathbf{446}$\textbf{K} & \textbf{6} & \textbf{25} & free-form & \textbf{395} & \textbf{345K} & open \& \textbf{cluttered}\\
\bottomrule
\end{tabular}
\begin{tablenotes}
\item[]$^\dag$RF-Pose is not publicly available.
\end{tablenotes}
\end{threeparttable}
}
\label{table: heatmap}
\end{table}

\section{Related Work: Indoor Radar Perception Datasets}

As shown in Table~\ref{table:datasets}, the last few years have witnessed the release of several indoor radar perception datasets. Most datasets focus on exploring single-view LR point clouds for tasks such as action classification and pose estimation. RadHAR~(\cite{RadHAR19}) collected a dataset consisting of point clouds using mmWave radar to classify only $5$ human activities from $2$ subjects. 
mm-Pose~(\cite{mmPose20}) used the CNN as the backbone network to estimate and track $25$ body joints when the subject performed one of the $4$ actions: walking, left-arm swing, right-arm swing, and both-arm swing. The dataset is rather limited with $39.7$K frames in total. mmMesh~(\cite{mmMesh21}) used single-view LR radar point clouds to directly estimate realistic-looking human meshes with encoded human body models to overcome the sparsity of point clouds. It collected radar data on $8$ daily activities from $20$ subjects within a confined open-space area, yielding $480$K radar frames in total. Both mm-Pose and mmMesh datasets are not publicly accessible. More recently, mRI~(\cite{mRI22}) and MM-Fi~(\cite{mmFi23}) targeted indoor rehabilitation activities and related pose estimation  ($14$ daily activities and $13$ rehabilitation exercises) from multi-modal sensors with single-view mmWave radar point cloud included. 

RF-Pose~(\cite{RFPose18}) and HuPR~(\cite{HuPR23}) are the two datasets leveraging two orthogonal radar heatmaps with an angular (azimuth and elevation) resolution of $15^\circ$. While RF-Pose was comprehensive in terms of data size and the diversity of subjects ($100$) and rooms ($50$), it is not publicly accessible. On the other hand, HuPR released $141$K data frames with 2D bounding boxes and keypoint annotations. However, as shown in Table~\ref{table: heatmap}, the diversity of rooms, subjects, and actions is rather limited and no segmentation annotations are included. In 2023, HIBER partially released $179$K data frames collected from $10$ subjects in a single room with an open foreground, using two orthogonal high-resolution radar heatmaps.  To bridge the gap between open-source HIBER and inaccessible RF-Pose, our dataset MMVR emphasizes the diversity of rooms, subjects and tasks, setting a stage for cross-environment and cross-subject performance evaluation and advancing the generalization capability of radar perception models.  Moreover, our dataset has almost double the size of annotation labels than that of HIBER.  

\begin{figure}[t]
    \centering
    \includegraphics[width=\textwidth]{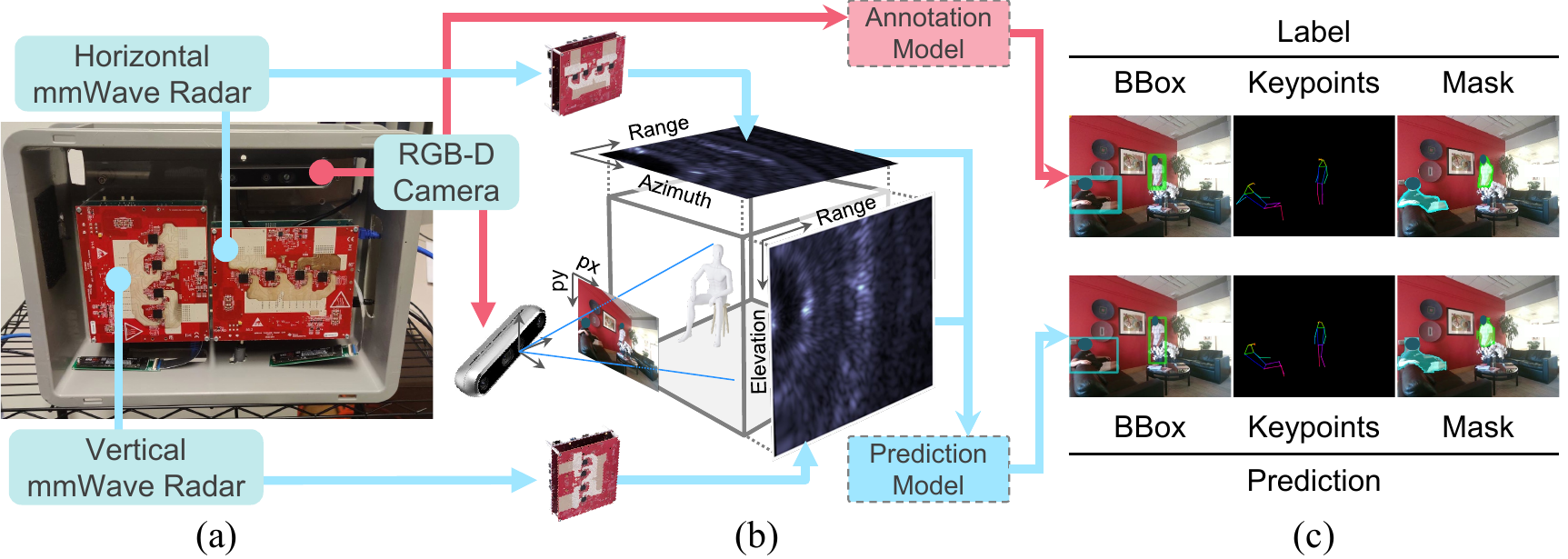} 
    \caption{MMVR sensor setup (a) with two mmWave radars and one RGB-D camera; (b) each frame includes two orthogonal high-resolution radar heatmaps and the synchronized RGB image; (c): the two radar heatmaps are inputs to radar perception models for the three perception tasks with the supervision from RGB-based labels. } 
    \label{fig:testbed}
\end{figure}

\section{MMVR Sensor Setup}
For MMVR, we develop a portable sensor testbed with two high-resolution radar sensors and a paired RGB-D camera in Fig.~\ref{fig:testbed}~(a). Following the red line, we use the Intel RealSense D455 sensor to obtain images of the scene in Fig.~\ref{fig:testbed}~(b) and generate image-based annotation labels in Fig.~\ref{fig:testbed}~(c) via pretrained annotation models. Separately, following the light blue  line,  we use two TI AWR2243 mmWave cascade radars to generate two radar views in the range-azimuth and range-elevation domains in Fig.~\ref{fig:testbed}~(b), and both views are fed to baseline prediction models to output prediction results for three perception tasks in Fig.~\ref{fig:testbed}~(c).   

\subsection{RGB-D Camera}
Thanks to the small form factor of the RealSense camera, we place it on the upper right side of our testbed box, just right above the horizontal radar sensor. RealSense camera D455 provides up to $1280 \times 800$ resolution for RGB images and up to $1280 \times 720$ resolution for stereo depth. The depth operating range is about \SI{6}{\meter} (varies with lighting
conditions). 
In our data collection, we choose the image resolution of $480 \times 640$ with a frame rate of $15$ fps.
As shown in Fig.~\ref{fig:testbed}~(b), the RGB image can be projected from the 3D camera coordinate into the image plane using a pinhole camera model.
The timestamp of RGB-D images is synchronized to the connected desktop with a sample timestamp accuracy of $50$ microseconds.

\subsection{Multi-View High-Resolution Radar}
\label{radarSensor}

\begin{wrapfigure}[11]{r}{2.6in}
    \vspace{-0.15in}
    \includegraphics[width=0.475\textwidth]{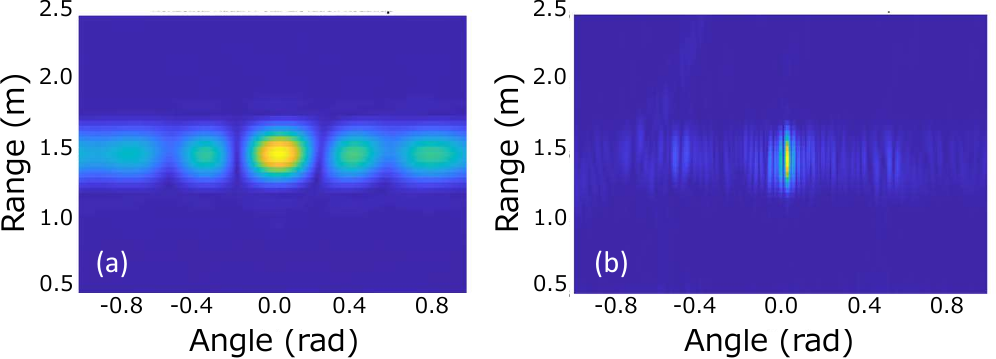}
    \caption{\label{fig:lrhr} Comparison between (a) low-resolution  and (b)  high-resolution radar heatmaps of a corner reflector.}
\end{wrapfigure}
To achieve high resolution in both azimuth and elevation domains, we place two AWR2243 cascade  radar sensors along the horizontal and vertical orientations in Fig.~\ref{fig:testbed}~(a). 
Coherently combining $4$ single-chip FMCW radar sensors, the cascade radar forms a virtual array of $86$ half-wavelength-spaced elements and offers an angular resolution of $1.3^\circ$ at the boresight direction\footnote{The boresight direction is the direction of peak gain of the antenna array.}, significantly better than the angular resolution of $15^\circ$ offered by the single-chip radar chips, e.g., IWR1443 and IWR1843, used by RadHAR~(\cite{RadHAR19}), mRI~(\cite{mRI22}), MM-Fi~(\cite{mmFi23}), and HuPR~(\cite{HuPR23}). With the perpendicular radar configuration, the two radars achieve simultaneously high-resolution radar views in both the range-azimuth and range-elevation planes, as illustrated in Fig.~\ref{fig:testbed}~(b). Fig.~\ref{fig:lrhr} compares the range-azimuth heatmap of  a corner reflector using the high-resolution (AWR2243) and low-resolution (IWR1843) radar sensors. 

\begin{figure}[b]
    \centering
    \includegraphics[width=\textwidth]{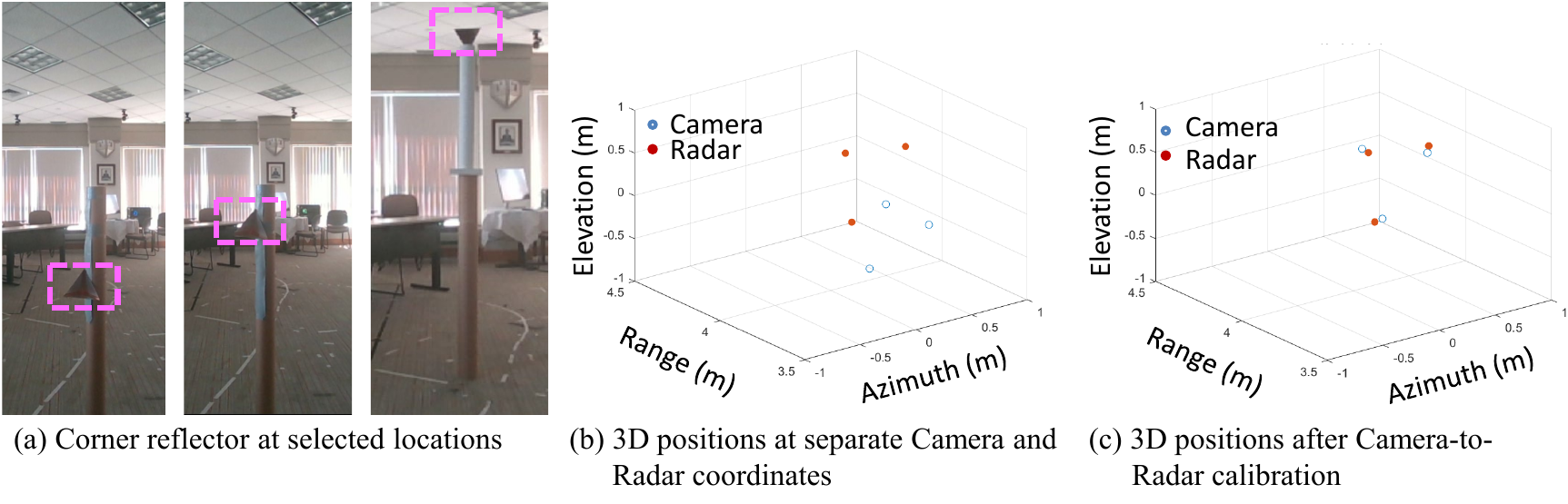} 
    \hfill 
    \caption{Camera-Radar coordinate calibration. } 
    \label{fig:calibration}
\end{figure}
\subsection{Calibration}
\label{calibration}
The camera-radar coordinate calibration is performed using measured 3D positions of a corner reflector at multiple locations in the camera and, respectively, radar coordinate systems. Denoting the measured camera-coordinate and radar-coordinate positions as $\mathbf{B}_{\text{camera}}, \mathbf{A}_{\text{radar}} \in \mathbb{R}^{3 \times N}$, where $N$ is the number of calibration positions, the radar-camera calibration is done by finding a rotation matrix $\mathbf{R}$ and a translation vector $\mathbf{t}$ such at 
\begin{align} \label{calib}
\min_{\mathbf{R}, \mathbf{t}} \sum_{i=1}^N\| \mathbf{R} \mathbf{B}_{\text{camera}}[:,i] + \mathbf{t} - \mathbf{B}_{\text{radar}}[:,i] \|^2_2
\end{align}
The above calibration can be solved using the singular value decomposition (SVD) to align two sets of points in the 3D space. Refer to Appendix~\ref{app:calibration} for the detailed calibration steps. Fig.~\ref{fig:calibration} shows the corner reflector at multiple calibration locations (a) and its positions in the camera and radar coordinate systems before (b) and after (c) the camera-radar coordinate calibration. 

\section{Dataset}
MMVR supports a range of perception tasks: object detection, pose estimation, and instance segmentation under a diverse setting of $6$ rooms and $25$ subjects and spanning over $9$ days.

\begin{figure}[b]
  \centering
        \includegraphics[width=\textwidth]{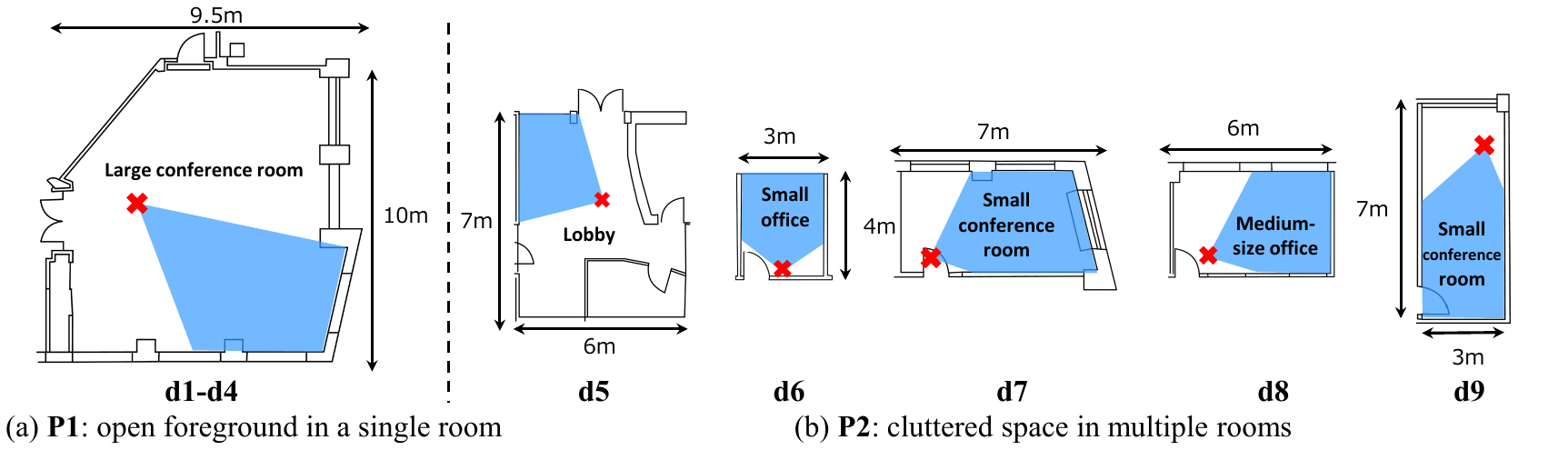} 
  \caption{Floorplans of $6$ diverse room settings for \textbf{Protocol 1} and  \textbf{Protocol 2}.} 
  \label{fig:floorplan_p1p2}
\end{figure}
\subsection{Data Collection}
MMVR consists of $345$K data frames collected in $2$ protocols: 
\begin{itemize}
    \item \textbf{Protocol 1 (P1: Open Foreground)}: $107.9$K data frames in a single open-foreground room with a single subject; see Fig.~\ref{fig:floorplan_p1p2}~(a) for the floorplan of \textbf{P1} and the first row of Fig.~\ref{fig:snapshots}~(c) for a snapshot. These data were collected over 4 separate days (d1-d4) with one or two sessions per day. The subject walking and jumping in the space remains unobstructed to both radar and RGB camera observations. 
    \item \textbf{Protocol 2 (P2: Cluttered Space)}: $237.8$K data frames in $5$ cluttered rooms with single and multiple subjects; see Fig.~\ref{fig:floorplan_p1p2}~(b) for the floorplans of \textbf{P2} and the second and third rows of Fig.~\ref{fig:snapshots}~(c) for snapshots. Starting from Day 5 (d5), $6$ data sessions were collected in one room. During the data collection, the subjects were doing diverse activities such as walking, sitting, stretching, reading, writing on the board, and having conversations. Additional snapshots for all data sessions are provided in Appendix~\ref{app:snapshot}. 
\end{itemize}

We split all data frames into non-overlapping $1$-min data segments, each having about $900$ data frames, given the frame rate of $15$ fps. Each data frame includes one RGB frame, one horizontal radar heatmap frame, one vertical radar heatmap frame, bounding box labels, segmentation labels, and keypoint labels. In total, we have $122$ data segments for \textbf{P1} and, respectively, $273$ data segments for \textbf{P2}. More details about the data collection protocol, number of segments in each data session, occlusion, and activities can be found in Table~\ref{tab:detailed_dataset_contents} of Appendix~\ref{app:dataset}. 

\begin{figure}[t]
    \centering
    \includegraphics[width=\textwidth]{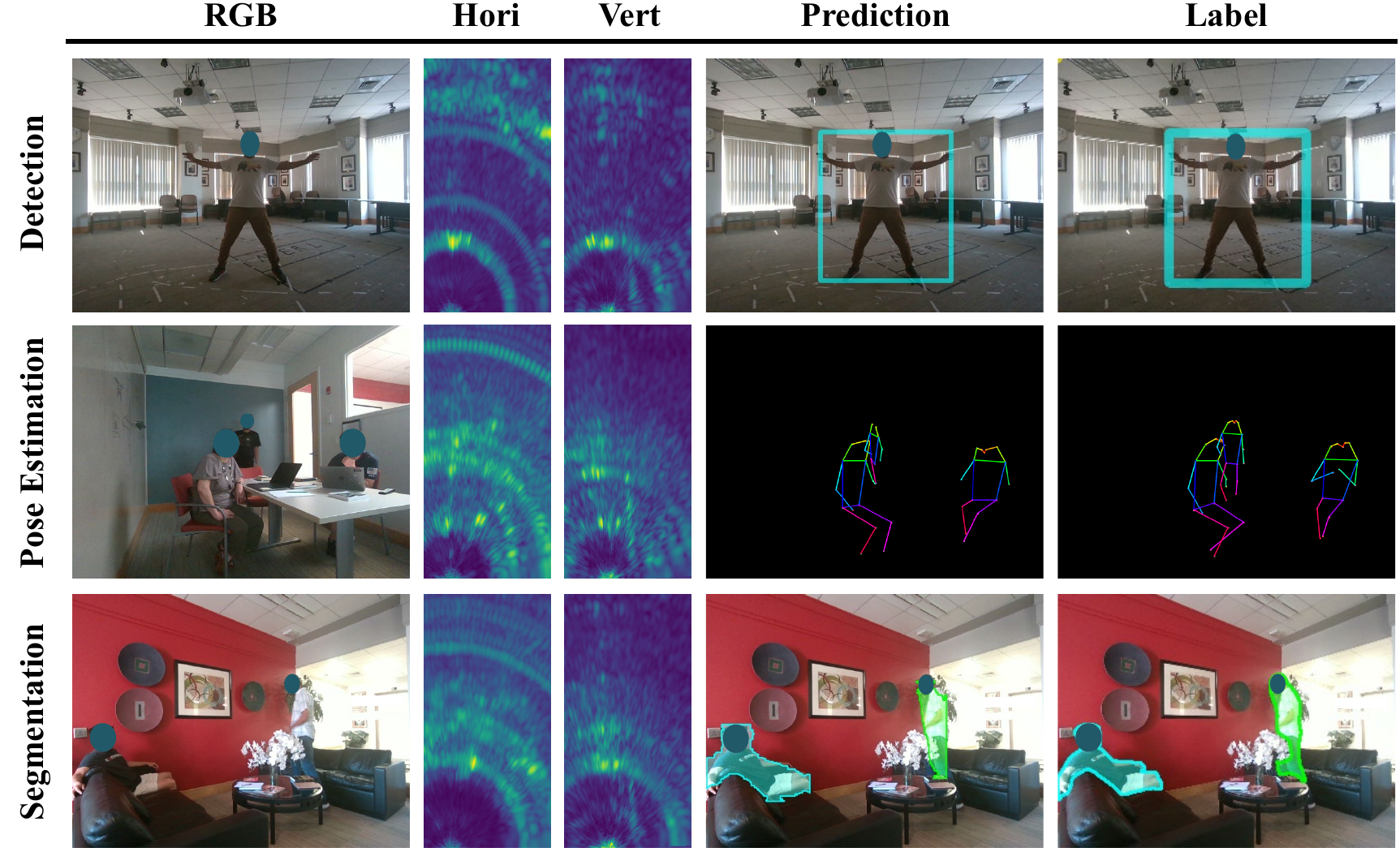} 
    \caption{Visualization of data collection (\textbf{P1: open foreground} in the first row and \textbf{P2: cluttered space} in the second and third rows), RGB (first column), corresponding multi-view (horizontal and vertical) radar heatmaps (second and third columns), baseline prediction results (fourth columns), and annotation labels (fifth columns).} 
    \label{fig:main_vis}
\end{figure}

\subsection{Annotation}
\label{sec:annotation}
We utilize pretrained image-based deep learning models: Mask2Former~(\cite{mask2former}) and HRNet~(\cite{hrnet}), to generate annotation labels. 

\subsubsection{Object Detection}
We take our RGB image through a pretrained backbone, i.e., the Swin Transformer, to get a list of low-resolution feature maps. Then Mask2Former~(\cite{mask2former, mmdetection}) enhances these features using a pixel decoder module to get high-resolution features. The Mask2Former decoder takes in a set of input queries, cross-attends the queries to feature maps from the encoder, and transforms them into a set of bounding boxes and classes. For our dataset, we only extract the bounding boxes with the ``person'' class with a shape of $(n,5)$ for $n$ objects and $[x_\text{min}, y_\text{min}, x_\text{max}, y_\text{max}, \text{confidence score}]$ for each object. One example is shown in the last column (Label) of the first row of Fig.~\ref{fig:main_vis}, where $n=1$. 

\subsubsection{Pose Estimation}
HRNet~(\cite{hrnet, mmpose2020}) maintains high-resolution representations by connecting high-to-low resolution convolutions in parallel, where there are repeated multiscale fusions across parallel convolutions. Such strong high-resolution representations can support multi-person semantic keypoint extraction. For each RGB frame, we applied a HRNet pretrained with the COCO Keypoints 2017 dataset to extract $17$ COCO keypoints for each person. The shape of keypoint labels is given as $(n,17,3)$ for $n$ objects and each keypoint represented by $3$ elements: (pixel$_\text{height}$, pixel$_\text{width}$, confidence score). 
One example is shown in the last column (Label) of the second row of Fig.~\ref{fig:main_vis}, where $n=3$.

\subsubsection{Instance Segmentation}
Mask2Former~(\cite{mask2former, mmdetection}) extends the above object detection to instance segmentation by further regressing the set of input queries into  a set of binary mask and class predictions, conditioned on the pixel decoder's features. The format of the instance segmentation is a $(n, 480, 640)$ binary (Boolean) tensor, where $480$ and $640$ are the height and width of the RGB image, respectively. 
One example is shown in the last column (Label) of the third row of Fig.~\ref{fig:main_vis}, where $n=2$ for two subjects in the RGB image.

\begin{figure}[t]
    \centering
    \includegraphics[width=\textwidth]{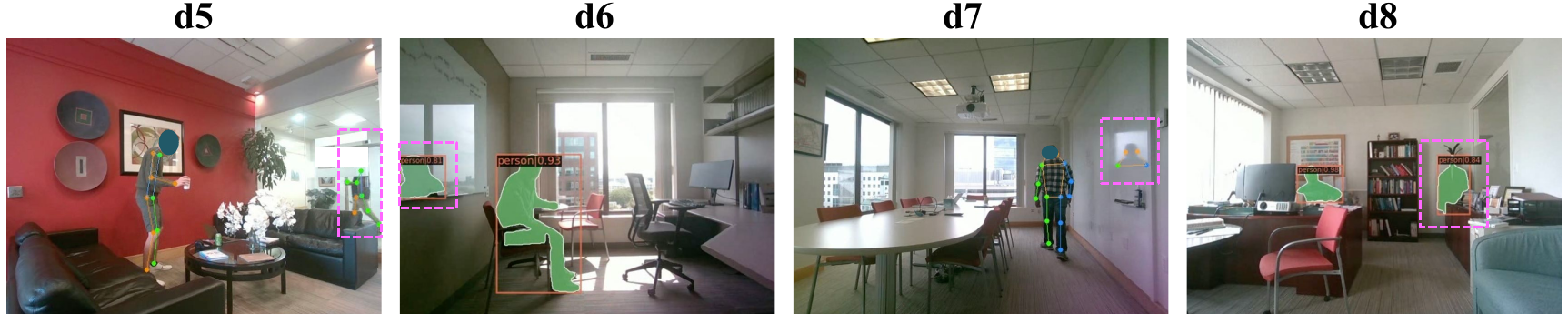} 
    \caption{Annotation curation: remove spurious labels (magenta boxes) due to the human presence behind the glass, highly reflective whiteboards, and a hanging suit. } 
    \label{fig:curation}
\end{figure}

\subsubsection{Annotation Curation}
\begin{wrapfigure}[7]{r}{2.7in}
    \centering
    \vspace{-0.3in}
    \includegraphics[width=0.48\textwidth]{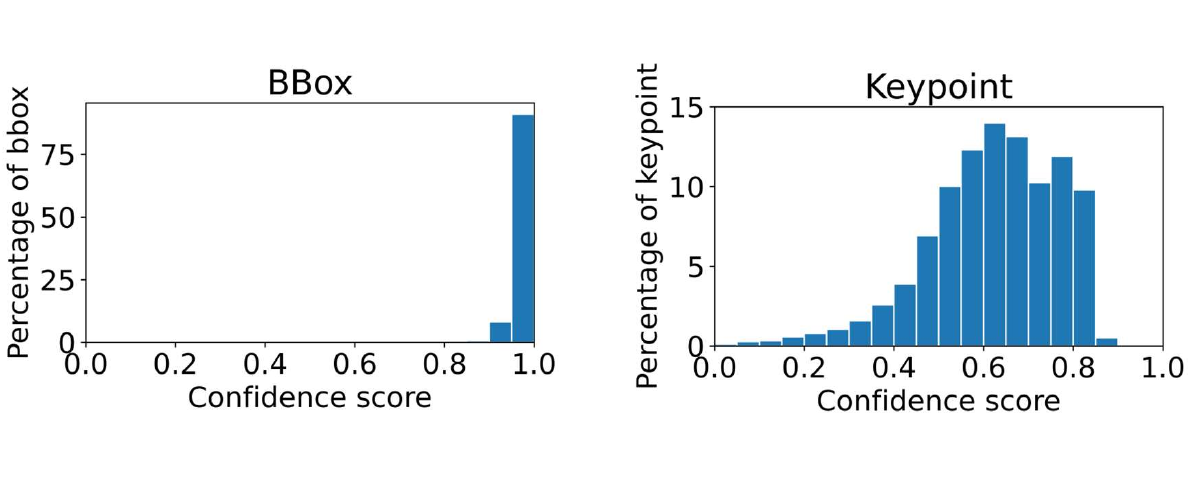}
    \vspace{-0.2in}
    \caption{\label{fig:score} Quality of curated annotations.}
\end{wrapfigure}
Although the above annotation models facilitate automatic label generation at a high volume, spurious labels still exist. 
For instance, in the lobby setting of d5 in Fig.~\ref{fig:curation}, RGB-based annotation pipelines generate high-confident annotation labels for passing-by people outside of the lobby (behind the glass), while radar sensors operating at \SI{77}{\giga\hertz} have limited penetration capability through the glass. As a result, we remove these spurious annotation labels from frames when there are people passing by.  One can find other cases in offices where a highly reflective whiteboard is present (see d6 and d7 of Fig.~\ref{fig:curation}) and where there is a hanging suit (see d8 of Fig.~\ref{fig:curation}). After data curation, Fig.~\ref{fig:score} shows the histograms of confidence scores of the annotated bounding box (BBox) and keypoints.
Please refer to Appendix~\ref{app:statistics} for other statistics of annotation in our dataset.

\subsection{Preprocessing for Multi-View Radar Heatmaps}

\begin{wrapfigure}[11]{r}{3.0in}
\vspace{-0.1in}
\centering
\includegraphics[width=0.525\textwidth]{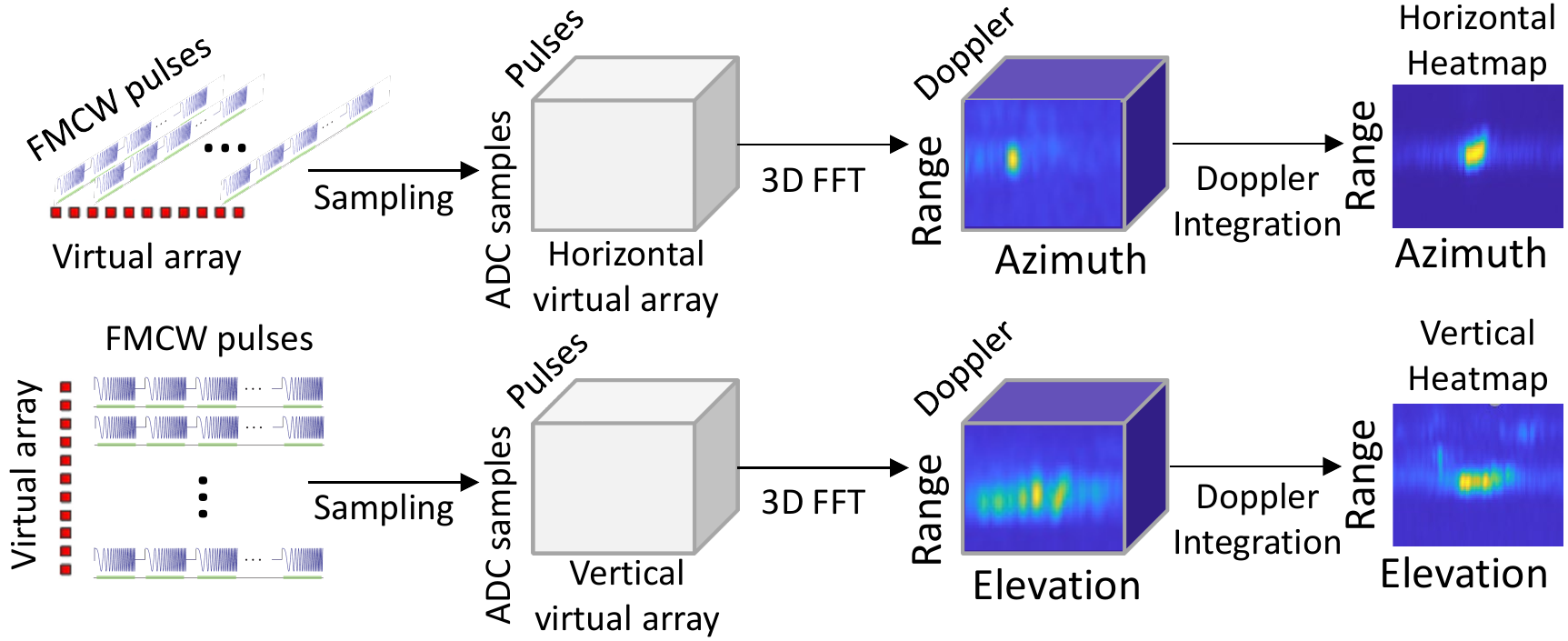}
\caption{\label{fig:radar} Multi-view heatmap preprocessing.}
\end{wrapfigure}

Fig.~\ref{fig:radar} shows our multi-view radar heatmap preprocessing chart from the two radar sensors forming two orthogonal virtual arrays of $86$ half-wavelength-spaced elements while sending multiple pulses.
By sampling the returned pulses, one can collect a 3D data cube along (horizontal/vertical) virtual array, ADC samples (intra-pulse or fast-time), and pulse (inter-pulse or slow-time) samples.
By applying 3D fast Fourier transform (FFT) over the datacube, one can obtain radar spectrum over the angle (azimuth for horizontal radar and elevation for vertical radar), range, and Doppler velocity domains. We further integrate the 3D radar spectrum along the Doppler domain to enhance the SNR, generating two (range-azimuth and range-elevation) radar heatmaps in the radar polar coordinate. We further project the two radar heatmaps into the radar Cartesian coordinate system and, with the help of the radar-camera calibration of Sec.~\ref{calibration}, convert them into range-azimuth and range-elevation views in the camera Cartesian coordinate system. 

\subsection{Synchronization}
For synchronization between radar and camera, we use two alignment steps: an initial alignment and a refined alignment. For the initial alignment, we place a metal corner reflector in front of our MMVR sensor board at the beginning of each session such that we can identify the starting frame for the horizontal radar sensor, vertical radar sensor, and RGB camera; see Appendix~\ref{app:sync} for an illustration. Even with the initial alignment, the clocks of the three sensors may drift over time. To this end, we perform a refined alignment to adjust the frames of all three sensors by synchronizing keyframes where notable motions, e.g., sit, stand up, are aligned at all three sensors. 

\section{Evaluation and Benchmarks}
\label{sec:eval}
In this section, we introduce the benchmark setup, performance metrics, baseline methods, and main results.

\subsection{Benchmark Setup}
\label{sec:setup}
To evaluate the model, we provide two data split strategies.
\begin{itemize}
    \item \textbf{Data Split 1 (S1: Random Split)} randomly splits the non-overlapping 1-min data segments ($122$ for \textbf{P1} and $273$ for \textbf{P2}) into train, validation and test sets at a ratio of $80:10:10$.
    \item \textbf{Data Split 2 (S2: Cross-Session and Unseen Split)} first splits all data segments in d5, d6, d7, and d9 into train, validation, and test sets. Then, we include all data in d8 in the test set such that one can assess the generalization performance of trained model for an unseen environment (d8).
\end{itemize}
The details about \textbf{S1} and \textbf{S2} can be found in Table~\ref{tab:detailed_dataset_splits_for_eval} of Appendix~\ref{app:datasettings}. 

\subsection{Metrics and Methods}
\label{sec:metric_and_method}

\subsubsection{Object Detection}
We adopt average precision on intersection over union~(IoU)~(\cite{Everingham2010_pascalvoc}) as an evaluation metric. 
IoU is the ratio of the overlap to the union of a predicted BBox $A$ and annotated BBox $B$ as:
\begin{equation}\label{eq:iou}
    \mathrm{IoU}(A, B) = \frac{|A \bigcap B|}{|A \bigcup B|}.
\end{equation}
We present three variants of average precision: AP$_{50}$, AP$_{75}$, and AP, where the former two represent the loose and strict constraints of IoU, while AP is the averaged score over $10$ different IoU thresholds in $[0.5, 0.95]$ with a stepsize of $0.05$. 
We use the latest RFMask~(\cite{RFMask23}) as the baseline method but modify it such that we can directly calculate the BBox loss using the BBox annotation in the image plane rather than the BBox loss at the two radar-view planes in HIBER. More details are described in Appendix~\ref{app:methods_prameters}.

\subsubsection{Pose Estimation}
We adopt average precision on object keypoint similarity~(OKS)~(\cite{lin2015microsoft}) as an evaluation metric.
OKS is scale-invariant and defined as
\begin{equation}\label{eq:oks}
    \mathrm{OKS}_i = \exp\left(-\frac{d_i^2}{2 s^2 k_i^2}\right), \quad i = 1, \cdots, 17, 
\end{equation}
for the $i$-th keypoint, where $d_i$ is the Euclidean distance between the prediction and corresponding ground truth, $k_i$ is a constant predefined uniquely for each joint, and $s$ is the scale of the subject being targeted.
Similarly, we calculate AP$_{50}$, AP$_{75}$, and AP for each keypoint and the average score over the $17$ keypoints. We re-implement RF-Pose~(\cite{RFPose18}) from  scratch for the baseline and further extend it to predict both keypoint-based heatmap and part affinity field~(PAF)~(\cite{OpenPose17}) for more robust multi-person association.

\subsubsection{Instance Segmentation}
Instance masks are predicted by RFMask, and average precision over IoU is used as the evaluation metric.
The segmentation IoU is computed by Eq.~\ref{eq:iou}, where the area of interest is determined by pixels, instead of BBoxes, that are predicted or annotated to belong to a person~(\cite{Everingham2010_pascalvoc}).

\subsection{Main Results}

\subsubsection{Object Detection}
Table~\ref{tab:main_results} shows the baseline performance of object detection from RFMask by taking $4$ consecutive radar frames as input.  It achieves AP$_{50}$ scores at more than $65$ under P1 for both data splits. These AP scores of our dataset MMVR are comparable to those numbers previously reported in the literature. For instance, RF-Pose~(\cite{RFPose18}) reported an AP score at $30.8$ with $6$ radar frames, which is similar to our benchmarks of AP at $25.53$ and $24.46$ under P1. Qualitatively, we observe more failure cases when the subject was about \SI{1}{\meter} or closer to the radar sensors.

For P2, the performance under S1 is much better than that of S2 as there is a significant drop in all AP metrics, suggesting limited generalization capability of the baseline method, particularly towards unseen cluttered environments. 

\begin{table}[t]
    \centering
    \footnotesize
    \caption{Baseline benchmarks for $3$ tasks under $2$ protocols and $2$ data splits.}
    \setlength\tabcolsep{8pt}
    \begin{tabular}{ccccccccccccc}
        \toprule
        \multirow{2}{*}{\textbf{Protocol}} & \multirow{2}{*}{\textbf{Split}} & \multicolumn{3}{c}{\textbf{Object Detection}} && \multicolumn{3}{c}{\textbf{Pose Estimation}} && \multicolumn{2}{c}{\textbf{Instance Seg.}} \\
        \cline{3-5} \cline{7-9} \cline{11-12}
        & & AP & AP$_{50}$ & AP$_{75}$ && AP & AP$_{50}$ & AP$_{75}$ && \multicolumn{2}{c}{IoU} \\
        \midrule
        \multirow{2}{*}{P1} & S1 & 25.53 & 67.30 & 15.86 && 46.24 & 62.88 & 47.45 && \multicolumn{2}{c}{71.98}\\
                            & S2 & 24.46 & 66.82 & 11.22 && 29.82 & 43.03 & 30.29 && \multicolumn{2}{c}{67.03} \\
        \midrule
        \multirow{2}{*}{P2} & S1 & 31.37 & 61.50 & 27.48 && 32.13 & 44.22 & 32.58 && \multicolumn{2}{c}{65.30}\\
                            & S2 & 6.03  & 22.77 & 0.88  && 7.11  & 11.98 & 6.76  && \multicolumn{2}{c}{56.07} \\
        \bottomrule
    \end{tabular}
    \label{tab:main_results}
\end{table}

\subsubsection{Pose Estimation}
\begin{figure*}[t]
    \centering
    \includegraphics[width=0.95\textwidth]{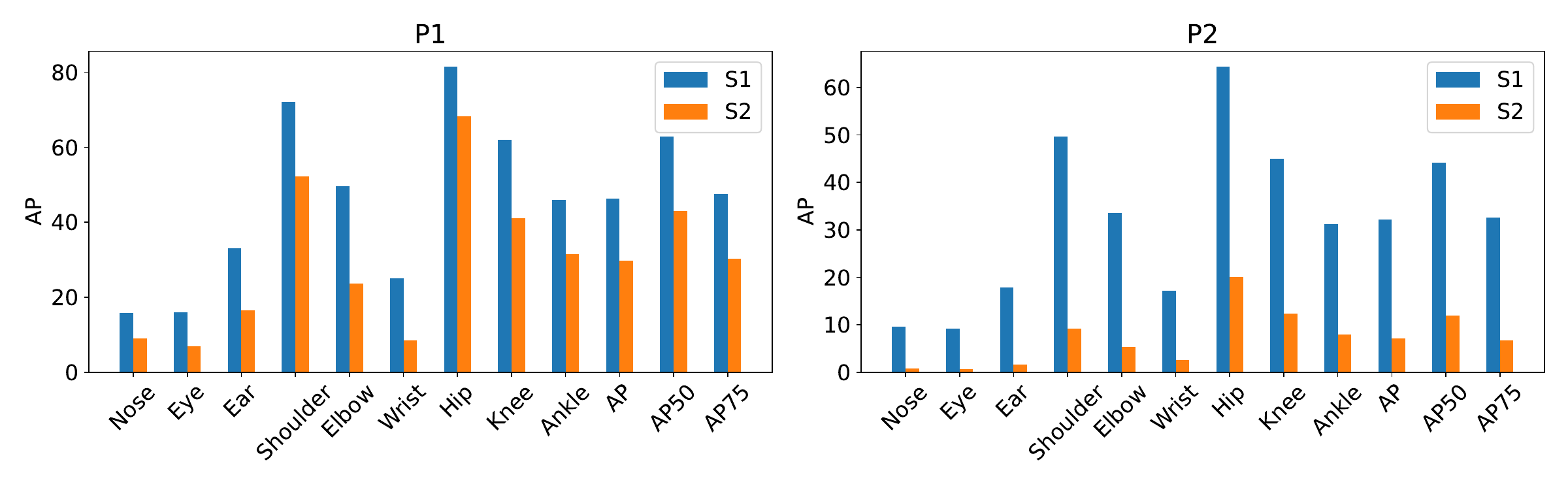}
    \hfill 
    \vspace{-0.1in}
    \caption{AP breakdown over keypoints under 2 protocols and 2 data splits. } 
    \label{fig:pose-split}
\end{figure*}

Table~\ref{tab:main_results} also consists of the baseline performance of pose estimation using RF-Pose with $4$ consecutive radar frames. Under P1, it achieves AP$_{50}$ of about $63$ and $43$ for S1 and S2, respectively. On the other hand, under P2, it maintains reasonable performance for S1 with an AP$_{50}$ score at $44.22$, while the performance drops significantly when the data split is S2, a similar effect we observed previously for object detection. Fig.~\ref{fig:pose-split} further breakdowns the AP score for multiple keypoints. It is seen that lower AP scores are more likely for keypoints such as limb and face parts. 

\subsubsection{Instance Segmentation}
In the last column of Table~\ref{tab:main_results}, we show relatively good and consistent performance of RFMask over the four combinations of $2$ protocols and $2$ data splits. On one hand, under the same protocol (P1 or P2), the performance drops slightly from S1 to S2, suggesting better generalization capability over unseen environments. On the other hand, performance degradation appears to be slightly larger over protocols than over splits. 
 
\subsubsection{Visualization}
We provide visualization results for all three tasks in the column of Prediction of Fig.~\ref{fig:main_vis}. In these snapshots, the considered baseline methods are able to distinguish radar signals from the subject from those from the background. Additional visualization results of good and failure cases under all $6$ rooms are included in Appendix~\ref{app:visualization}. 

\subsection{Ablation Study}
\label{sec:ablation}
We report on several ablation studies using \textbf{S1} under \textbf{P1}. 

\subsubsection{Number of Frames}
From Table~\ref{tab:abl_num_frames_det_seg}, longer sequences generally lead to better performance for each task. In particular, the object detection performance is increased significantly when $12$ radar frames are used.
At a frame rate of $15$ fps,  $18$ frames for a duration of $1.2$ seconds lead to slightly worse performance than the use of $12$ frames. 

\begin{table}[t]
    \centering
    \footnotesize
    \caption{\textbf{\# of Frames}: Using longer time horizon improves the performance.}
    \setlength\tabcolsep{8pt}
    \begin{tabular}{cccccccccccc}
        \toprule
        \multirow{2}{*}{\textbf{\# of frames}} & \multicolumn{3}{c}{\textbf{Detection}} && \multicolumn{3}{c}{\textbf{Pose Estimation}} && \multicolumn{2}{c}{\textbf{Segmentation}} \\
        \cline{2-4} \cline{6-8} \cline{10-11}
          & AP & AP$_{50}$ & AP$_{75}$ && AP & AP$_{50}$ & AP$_{75}$ && \multicolumn{2}{c}{IoU} \\
        \midrule
        4 & 25.53 & 67.30 & 15.86 && 46.24 & 62.88 & 47.45 && \multicolumn{2}{c}{71.98}\\
        12 & \textbf{33.12} & 71.17 & \textbf{26.84} && \textbf{46.86} & \textbf{63.22} & \textbf{48.14} && \multicolumn{2}{c}{71.81} \\
        18 & 32.78 & \textbf{72.03} & 25.44 && 46.70    & 63.07    & 47.95 && \multicolumn{2}{c}{\textbf{73.28}}\\
        \bottomrule
    \end{tabular}
    \label{tab:abl_num_frames_det_seg}
\end{table}

\subsubsection{Single View versus Dual View}
Table~\ref{tab:abl_single_dual} presents a comparison between using a single view (feeding the horizontal view only), and using dual views (feeding both the horizontal and vertical views).
The results indicate that using the dual views improves performance for both RFMask and RF-Pose. 
This improvement implies that, while the horizontal view is informative enough to determine the location and relative scale of the subjects in the image plane, the vertical view may provide additional features about a subject in the elevation domain and add more fine-grained and distinguished characteristics for different body parts.

\subsubsection{Comparison of Detection Backbones}
We also take the more advanced DETR architecture~(\cite{Carion2020_detr}) as another simple baseline method without significant customization towards radar frames. We compare the performance of object detection and instance segmentation between RFMask and DETR in Table~\ref{tab:abl_comparison_method_det_seg}.
It appears that exploiting backbone feature self-attention and query-feature cross-attention in DETR leads to consistently better performance for the two considered tasks. 

\begin{table}[t]
    \begin{tabular}{cc}
        \begin{minipage}[t]{0.48\hsize}
            \footnotesize
            \caption{\textbf{Single/Dual}: Using the dual views works the best.}
            \setlength\tabcolsep{5pt}
            \begin{tabular}{cccccc}
                \toprule
                \textbf{Method} & \textbf{View} & \textbf{AP} & \textbf{AP$_{50}$} & \textbf{AP$_{75}$} & \textbf{IoU} \\
                 \midrule
                 \multirow{2}{*}{RFMask}
                 & Single & 24.64 & 66.17 & 15.21 & 70.64 \\
                 & Dual   & \textbf{25.53} & \textbf{67.30} & \textbf{15.86} & \textbf{71.98} \\
                 \midrule
                 \multirow{2}{*}{RF-Pose} 
                 & Single & 22.22 & 30.99 & 22.64 & -\\
                 & Dual   & \textbf{46.24} & \textbf{62.88} & \textbf{47.45} & -\\
                \bottomrule
            \end{tabular}
            \label{tab:abl_single_dual}
        \end{minipage} &
        \hspace{0.5em}
        \begin{minipage}[t]{0.43\hsize}
            \centering
            \footnotesize
            \caption{\textbf{Method}: DETR can improve the performance.}
            \setlength\tabcolsep{3.5pt}
            \begin{tabular}{ccccccccccc}
                \toprule
                \textbf{Method} & \textbf{View} & \textbf{AP} & \textbf{AP$_{50}$} & \textbf{AP$_{75}$} & \textbf{IoU} \\
                \midrule
                RFMask & Single & 24.64 & 66.17 & 15.21 & 70.64 \\
                DETR & Single & \textbf{35.78} & \textbf{78.59} & \textbf{28.02} & \textbf{72.65} \\
                \bottomrule
            \end{tabular}
            \label{tab:abl_comparison_method_det_seg}
        \end{minipage}
    \end{tabular}
\end{table}

\begin{table}[t]
    \centering
    \footnotesize
    \caption{Baseline benchmarks between HIBER \cite{RFMask23} and MMVR.}
    \setlength\tabcolsep{8pt}
    \resizebox{0.9\columnwidth}{!}{%
    \begin{tabular}{cccccccccc}
        \toprule
        \multirow{2}{*}{\textbf{Room Setting}} & \multirow{2}{*}{\textbf{Dataset}} & \multicolumn{3}{c}{\textbf{Object Detection}} && \multicolumn{2}{c}{\textbf{Instance Seg.}} \\
        \cline{3-5} \cline{7-8}
        & & AP & AP$_{50}$ & AP$_{75}$  && \multicolumn{2}{c}{IoU} \\
        \midrule
        {Open } & HIBER (Walk) & 17.77 & 52.46 & 6.78 && \multicolumn{2}{c}{48.47}\\
        % {Open } & HIBER (Walk except env1) & 54.95 & 98.89 & 51.64 && 28.89 & 63.55 & 22.07 && \multicolumn{2}{c}{49.03}\\
        {Foreground}  & MMVR (P1) & 24.46 & 66.82 & 11.22 && \multicolumn{2}{c}{67.03} \\
        \bottomrule
    \end{tabular}
    }
    \label{tab:dataset_compare}
    \vspace{-0.1in}
\end{table}

\subsubsection{Comparison over Different Radar Datasets}
An interesting ablation study involves training the radar perception model on one radar dataset, such as MMVR, and evaluating its performance on another dataset, like HIBER, or vice versa. However, this study requires \emph{dataset-dependent} 3D radar-camera coordinate calibration and 3D-to-2D projection in the camera coordinate, due to varying relative geometries between the camera and radar and the use of different cameras in different datasets. Instead, we directly compare the same tasks across different datasets under similar environmental settings: a single subject walking in an open-foreground room.  This is achieved using the ``walk'' data split in HIBER and  ``P1'' in our MMVR. As shown in Table~\ref{tab:dataset_compare}, the performance between the two datasets is comparable. Note that we used refined bounding boxes in the 2D image plane for the HIBER evaluation as the original bounding boxes were excessively large around the subjects.

\section{Responsibility to Human Subjects}
\label{sec:responsibility}
Our data collection for MMVR was approved by our institutional review board (IRB). We initiated a call-for-volunteers in our institution and recruited $25$ participants. At the beginning of each data session, we informed participants about our experiment, its research goal, the procedure,  potential exposure to high-frequency radio frequency waves, and the use of the camera. We notified the participants that de-identified data would be made publicly accessible for research purposes. A consent form was signed by all participants. 

\section{Limitations and Potential Harms}
\label{sec:limits}
Our dataset MMVR has limitations regarding annotations and benchmarks. Our annotation pipeline is based on RGB images. In cases of natural occlusion (chair, table, sofa) in a cluttered room of \textbf{P2}, the annotated BBox, keypoints, and segmentation pixels are also occluded or have extremely low reliability. As a result, it is unclear if radar-based approaches can lead to better perception performance than camera-based approaches under these natural occlusions. The maximum number of subjects in a session is limited to $3$. Radar perception in a dense crowd is less explored and the dataset is extremely limited. Although radar perception has fewer privacy concerns than the camera, MMVR can be potentially utilized to classify and estimate attributes of subjects such as gender, size, height, and gait. It may be also used to advance technologies for indoor surveillance without acknowledgment or permission. 

\section{Conclusion}
In this paper, MMVR scales up indoor radar data collection using multi-view high-resolution heatmaps in a multi-day, multi-room, and multi-subject setting, with an emphasis on the diversity of the environment and subjects. It complies with an extensive collection of $345$K radar heatmap frames from  $25$ subjects and $6$ different rooms to benchmark three mainstream perception tasks such as object detection, pose estimation, and instance segmentation with $446$K bounding boxes/segmentation instances and $7.59$ million keypoints annotated.  We hope that MMVR can stimulate the development of robust radar perception models that can operate effectively in varied real-world applications in smart home systems, security, elderly care, and navigation assistance for visually impaired individuals, contributing positively to society.

\section{Contributions}
M. Rahman, P. Wang, and P. Boufounos contributed to the development of the MMVR testbed. M. Rahman and P. Wang coordinated and participated in data collection. P. Li and P. Wang established the annotation pipeline for labels used in object detection, pose estimation, and segmentation with P. Wang contributing to label curation. S. Kato and P. Wang worked on the pose estimation baseline with S. Kato evaluating benchmarks under various data splits and protocols and visualizing the results. R. Yataka, A. Cardace, and P. Wang contributed to the object detection and segmentation baselines with R. Yataka evaluating benchmarks under various data splits and protocols benchmarks and visualizing the results. M. Rahman and P. Wang worked on the initial alignment between the camera and radar frames, the camera-radar calibration, and the signal processing pipeline to generate radar heatmaps. R. Yataka contributed to the refined alignment between camera and radar frames. S. Kato, R. Yataka, and P. Wang contributed to data segmentation, sequence splits, and the organization of the final data structure. P. Wang initiated the MMVR effort for indoor perception applications at MERL and led the project from its inception. P. Wang and P. Boufounos supervised the project.

%%%%%%%%%%%%%%%%%%%%%%%%%%%%%%%%%%%%%%%%%%%%%%%%%%%%%%%%%%%%%%%%%%%%%%%%%%%%%%%%%%%%%%%%%%%%%%%%%%%%%%%%%%%%%%%%
\clearpage
\bibliography{neurips_url}
\bibliographystyle{neurips}

%%%%%%%%%%%%%%%%%%%%%%%%%%%%%%%%%%%%%%%%%%%%%%%%%%%%%%%%%%%%
\clearpage
\appendix

%%%%%%%%%%%%%%%%%%%%%%%%%%%%%%%%%%%%%%%%%%%%%%%%%%%%%%%%%%%%%%%%%%%%%%%%%%%%%%%%%%%%%%%%%%%%%%%%%%%%%%%%%%%%%%%%
\section{Detailed Steps for Camera-Radar Calibration}
\label{app:calibration}
Given $N$ pairs of measured camera-coordinate positions $\mathbf{B}_{\text{camera}} \in \mathbb{R}^{3 \times N}$ and radar-coordinate positions $\mathbf{A}_{\text{radar}} \in \mathbb{R}^{3 \times N}$, the rotation matrix $\mathbf{R}$ and a translation vector $\mathbf{t}$ can be found by minimizing the Euclidean distance error in Eq.~\ref{calib}. This can be numerically solved in the following steps:
\begin{itemize}
    \item \textbf{Centering the Calibration Position Sets}: Compute the centroids of both position sets $\mathbf{A}_{\text{radar}}$ and $\mathbf{B}_{\text{camera}}$ as $\mubf_{\text{radar}} = \text{mean}(\mathbf{A}_{\text{radar}}, axis=1) \in \mathbb{R}^{3 \times 1}$ and $\mubf_{\text{camera}} = \text{mean}(\mathbf{B}_{\text{camera}}, axis=1) \in \mathbb{R}^{3 \times 1}$. Then translate both sets to have their centroids at the origin:
    \begin{align}
    \tilde{\mathbf{A}}_{\text{radar}} & = \mathbf{A}_{\text{radar}} -\mubf_{\text{radar}}, \notag \\
    \tilde{\mathbf{B}}_{\text{camera}} & = \mathbf{B}_{\text{camera}} -\mubf_{\text{camera}}.
    \end{align}
    This initial alignment separates the calculation of the rotation matrix from that of the translation.
    \item \textbf{Computing the Covariance Matrix}: Calculate the covariance matrix between the centered calibration position sets $\tilde{\mathbf{A}}_{\text{radar}}$ and $\tilde{\mathbf{B}}_{\text{camera}}$: 
    \begin{align}
   \Hbf  = \tilde{\mathbf{A}}_{\text{radar}} \tilde{\mathbf{B}}^T_{\text{camera}},
    \end{align}
    where $(\cdot)^T$ denotes the matrix transpose. 
    \item \textbf{Singular Value Decomposition (SVD)}: Perform the singular value decomposition on the covariance matrix:
    \begin{align}
   [\Ubf, \Sbf, \Vbf]  = \text{SVD}(\Hbf),
    \end{align}
    where $\Ubf$ and $\Vbf$ are unitary matrices, and $\Sbf$ is a diagonal matrix.
    \item \textbf{Determining the Rotation Matrix $\Rbf$}:  The rotation matrix is found as 
    \begin{align} \label{rotation}
    {\Rbf} = \Vbf \Ubf^T.
    \end{align}
    Since $\Ubf$ and $\Vbf$ are unitary matrices, $\Rbf$ is guaranteed to be a rotation matrix with a determinant of $1$. 
    \item \textbf{Calculating the Translation Vector $\tbf$}: Once the rotation matrix ${\Rbf}$ is found in Eq.~\ref{rotation}, the translation vector $\tbf$ can be determined as 
    \begin{align}
    \tbf = \mubf_{\text{camera}} - \Rbf \mubf_{\text{radar}}.
    \end{align}
    The above step first applies the rotation to the radar-coordinate calibration set centroid $\mubf_{\text{radar}}$ and then subtracts it from the camera-coordinate calibration set centroid $\mubf_{\text{camera}}$. 
\end{itemize}

\begin{figure}[t]
    \centering
    \includegraphics[width=\textwidth]{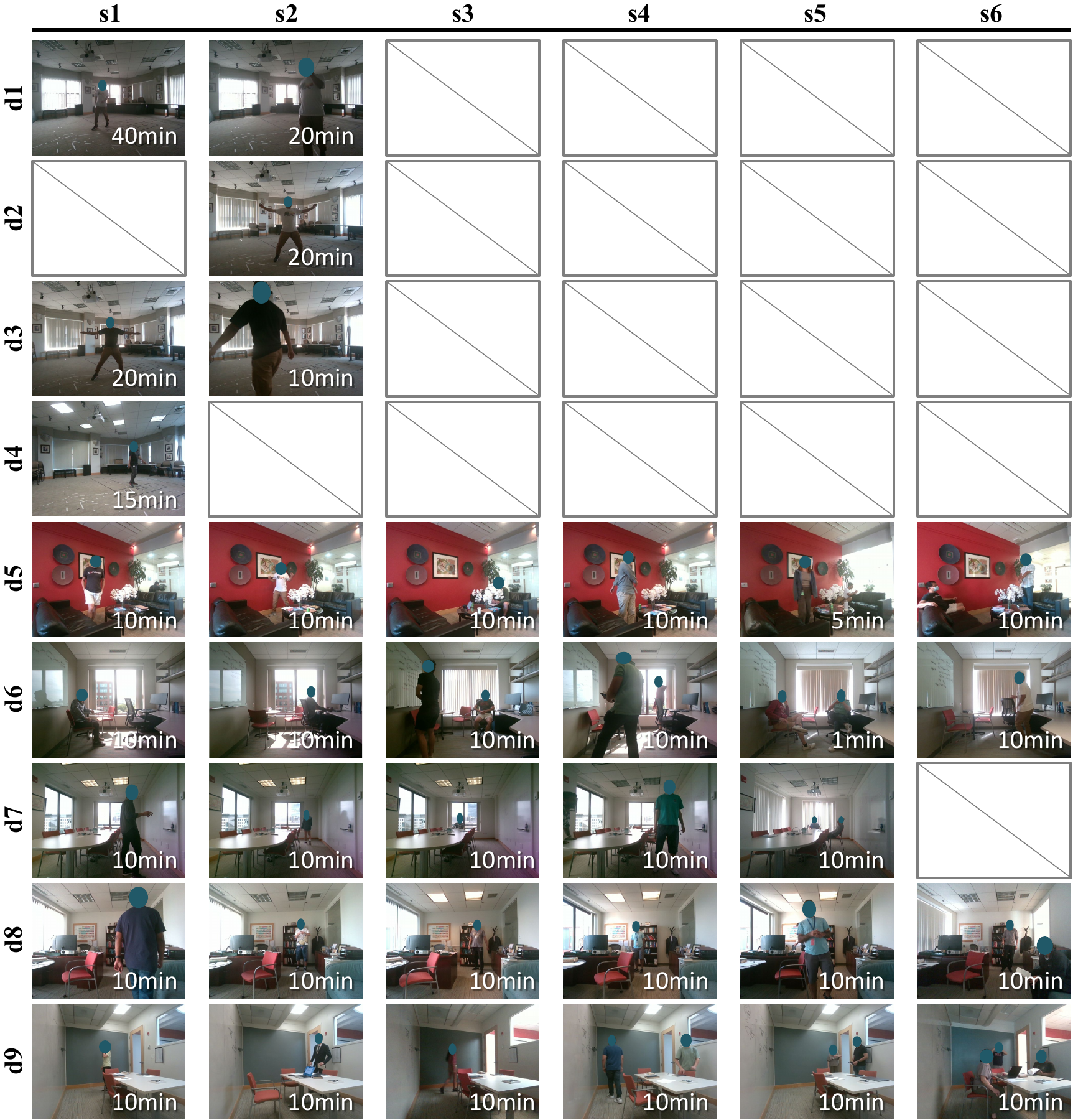}
    \caption{
        Gallery of additional snapshots.
    }
    \label{fig:additional_snapshots}
    \vspace{-0.1in}
\end{figure}

\section{Additional Snapshots of All Data Sessions}
\label{app:snapshot}
We provide additional snapshots of all data sessions in Fig.~\ref{fig:additional_snapshots}. Each row represents a day (\textbf{d}) while each column represents a session (\textbf{s}) with the session duration indicated.

\begin{table}[t]
    \centering
    \caption{Details of data sessions.}
    \setlength\tabcolsep{4pt}
    \resizebox{\columnwidth}{!}{ %
    \begin{tabular}{c|c|cc|ccc|l}
        \toprule
        \textbf{Protocol} & \textbf{Session} & \textbf{\# of seg.} & \textbf{\# of frames} & \textbf{Room} & \textbf{Occlusion} & \textbf{\# of Sbj.} & \multicolumn{1}{c}{\textbf{Action}} \\
        \midrule
        \multirow{7}{*}{\textbf{P1}}
        & d1s1 & 39 & 35,085 & large conference & - & 1 & walking \\
        & d1s2 & 20 & 17,618 & large conference & - & 1 & walking \\
        & d2s2 & 20 & 17,563 & large conference & - & 1 & jumping \\
        & d3s1 & 19 & 16,962 & large conference & - & 1 & jumping \\
        & d3s2 & 9 & 7,436 & large conference & - & 1 & walking \\
        & d4s1 & 15 & 13,238 & large conference & - & 1 & walking \\
        \cline{2-8}
        & Total d1-d4 & 122 & 107,902 & - & - & - & \multicolumn{1}{c}{-} \\
        \midrule
        \multirow{30}{*}{\textbf{P2}}
        & d5s1 & 9 & 7,808 & lobby & chair, table, vase & 1 & walking, sitting, stretching, reading \\
        & d5s2 & 9 & 7,407 & lobby & chair, table, vase & 1 & walking, sitting, reading, writing, stretching \\
        & d5s3 & 10 & 8,817 & lobby & chair, table, vase & 1 & walking, sitting, reading, writing, stretching, eating, drinking \\
        & d5s4 & 9 & 8,040 & lobby & chair, table, vase & 1 & walking, sitting, reading, writing, stretching \\
        & d5s5 & 6 & 5,382 & lobby & chair, table, vase & 2 & walking, sitting, reading, drinking, stretching, talking \\
        & d5s6 & 10 & 8,821 & lobby & chair, table, vase & 3 & walking, sitting, reading, stretching, talking \\
        & d6s1 & 10 & 8,592 & small office & chair & 1 & walking, sitting, reading, writing, stretching \\
        & d6s2 & 10 & 8,852 & small office & chair & 1 & walking, sitting, writing, stretching \\
        & d6s3 & 10 & 8,552 & small office & chair & 2 & walking, sitting, reading, writing, stretching, talking \\
        & d6s4 & 10 & 8,842 & small office & chair & 2 & walking, sitting, reading, writing, stretching, talking \\
        & d6s5 & 1 & 430 & small office & chair & 2 & walking, sitting, reading, writing, stretching, talking \\
        & d6s6 & 10 & 8,677 & small office & chair & 1 & walking, sitting, reading, writing, stretching \\
        & d7s1 & 10 & 8,757 & small conference & chair, table & 1 & walking, sitting, reading, writing, stretching \\
        & d7s2 & 10 & 8,392 & small conference & chair, table & 1 & walking, sitting, reading, writing, stretching \\
        & d7s3 & 10 & 8,855 & small conference & chair, table & 1 & walking, sitting, reading, writing, stretching \\
        & d7s4 & 10 & 8,844 & small conference & chair, table & 2 & walking, sitting, reading, writing, stretching, talking \\
        & d7s5 & 10 & 8,795 & small conference & chair, table & 2 & walking, sitting, reading, writing, stretching, talking \\
        & d8s1 & 10 & 8,623 & medium office & chair, table & 1 & walking, sitting, reading, stretching, talking \\
        & d8s2 & 10 & 8,177 & medium office & chair, table & 1 & walking, sitting, reading, writing, stretching, playing \\
        & d8s3 & 10 & 8,758 & medium office & chair, table & 1 & walking, sitting, reading, writing, stretching \\
        & d8s4 & 10 & 8,872 & medium office & chair, table & 1 & walking, sitting, reading, writing, stretching \\
        & d8s5 & 10 & 8,880 & medium office & chair, table & 2 & walking, sitting, reading, writing, stretching, talking \\
        & d8s6 & 10 & 8,805 & medium office & chair, table & 2 & walking, sitting, reading, writing, stretching, talking, playing \\
        & d9s1 & 9 & 7,964 & small conference & chair, table & 1 & walking, sitting, reading, writing, stretching \\
        & d9s2 & 10 & 8,795 & small conference & chair, table & 1 & walking, sitting, reading, writing, stretching \\
        & d9s3 & 10 & 8,770 & small conference & chair, table & 1 & walking, sitting, reading, writing, stretching \\
        & d9s4 & 10 & 8,677 & small conference & chair, table & 2 & walking, sitting, reading, writing, stretching, talking \\
        & d9s5 & 10 & 8,817 & small conference & chair, table & 2 & walking, sitting, reading, writing, stretching, talking \\
        & d9s6 & 10 & 8,797 & small conference & chair, table & 3 & walking, sitting, reading, writing, stretching, talking \\
        % \midrule
        \cline{2-8}
        & Total d5-d9 & 273 & 237,798 & - & - & - & \multicolumn{1}{c}{-} \\
        \midrule
        \multicolumn{2}{c|}{\textbf{Total}} & \textbf{395} & \textbf{345,700} & - & - & - & \multicolumn{1}{c}{-} \\
        \bottomrule
    \end{tabular}
    \label{tab:detailed_dataset_contents}    }
\end{table}

\begin{figure}[t]
    \centering
    \includegraphics[width=0.9\textwidth]{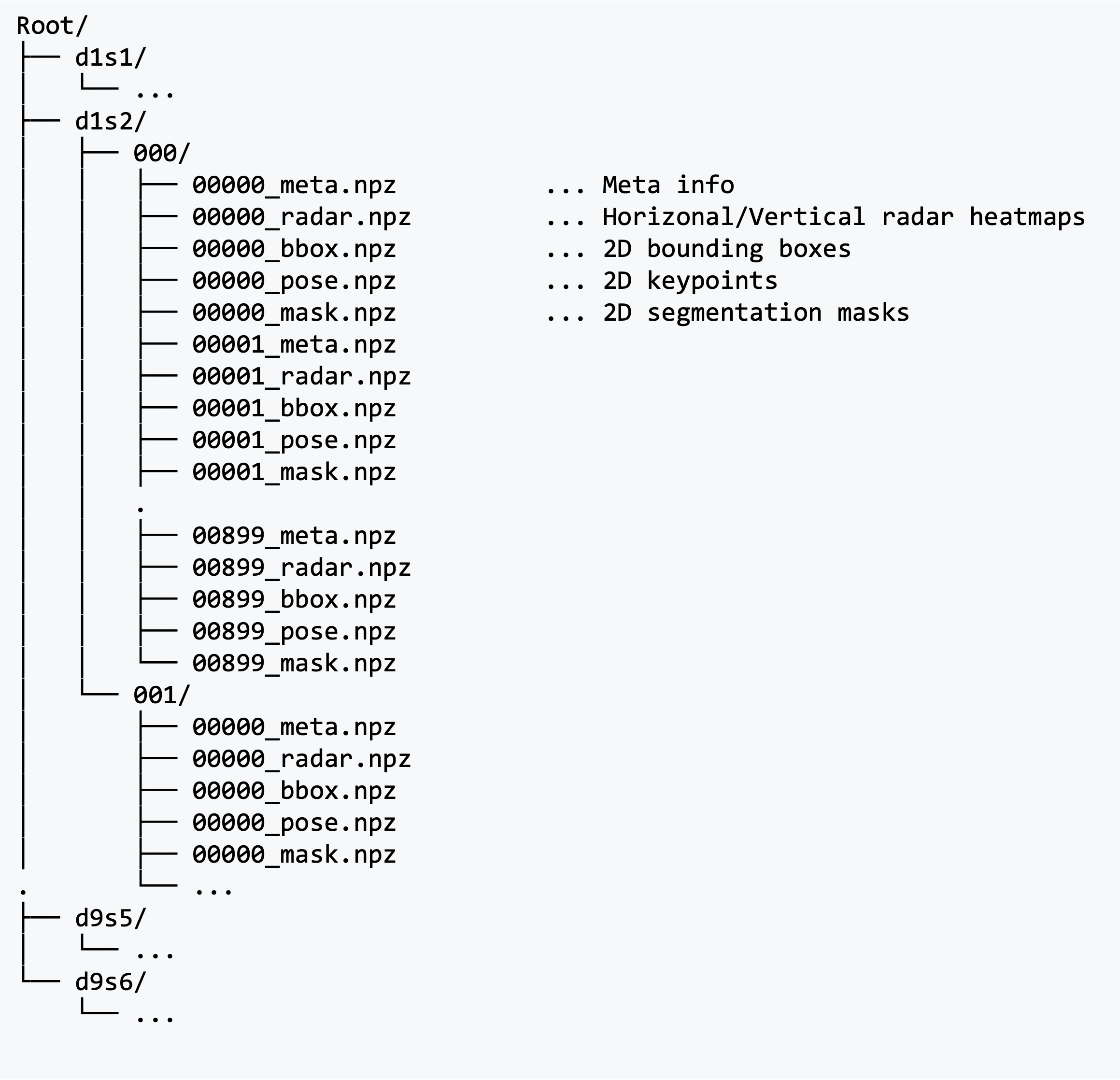} 
    \caption{
        The folder structure of our MMVR dataset. 
    }
    \label{fig:vis_data_structure}
\end{figure}

\section{Details of Data Sessions}
\label{app:dataset}

\subsection{Content}
Table~\ref{tab:detailed_dataset_contents} enumerates the details of all data sessions included in our dataset MMVR. Our dataset includes two protocols: \textbf{P1} and \textbf{P2}. Under each protocol, each session is split into non-overlapping one-minute data segments. The number of data segments in each session is listed in the column of \texttt{\# of seg.}, along with the column of \texttt{\# of frames}. Other statistics of each data session such as storage size, room type, obstacles for occlusion, number of subjects, and action types of the subjects are also listed. 

\subsection{Folder Structure}
Fig.~\ref{fig:vis_data_structure} illustrates the folder structure of our dataset MMVR. Below the root folder, we have one data folder of $\text{d}x\text{s}y$ for each data session, where $x$ and $y$ are the indices for the day and session, respectively. Under each data folder or data session, we group the data frames into separate non-overlapping data segment folders named using a three-digit, zero-filled convention based on the chronological order of the data segment. For instance, the first one-minute data segment corresponds to the folder $000$, while the second data segment folder is named after $001$. 

Under each data segment folder, there are three types of files: meta, radar, and annotation including bounding boxes, keypoints and segmentation masks for each radar frame, named using a five-digit, zero-filled convention. 
For instance, $\text{d}1\text{s}2/000/00001\_\text{meta}.\text{npz}$ contains the metadata for the second data frame $00001$ in the first data segment $000$ of the data session $\text{d}1\text{s}2$. Under the same data segment folder, $00001\_\text{radar}.\text{npz}$ contains the horizontal and vertical radar heatmaps. The annotation files are saved separately for each perception task to facilitate the data loading. For instance, $00001\_\text{bbox/pose/mask}.\text{npz}$ contain the bounding boxes, keypoints, and segmentation masks for the corresponding frame. 

\begin{figure}[t]
    \centering
    \begin{minipage}[b]{0.4\hsize}
        \centering
        \includegraphics[width=1.0\textwidth]{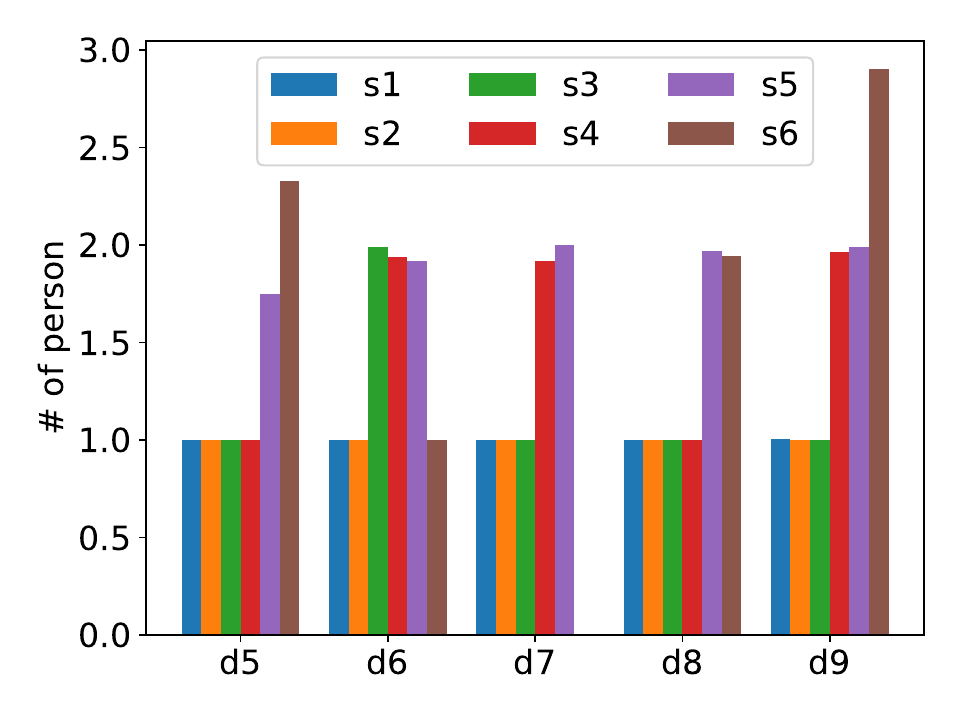}
        \caption{
            The number of labels per data session.
        }
        \label{fig:num_annotation_per_sess}        
    \end{minipage}
    \hspace{2em}
    \begin{minipage}[b]{0.4\hsize}
        \centering
        \includegraphics[width=1\textwidth]{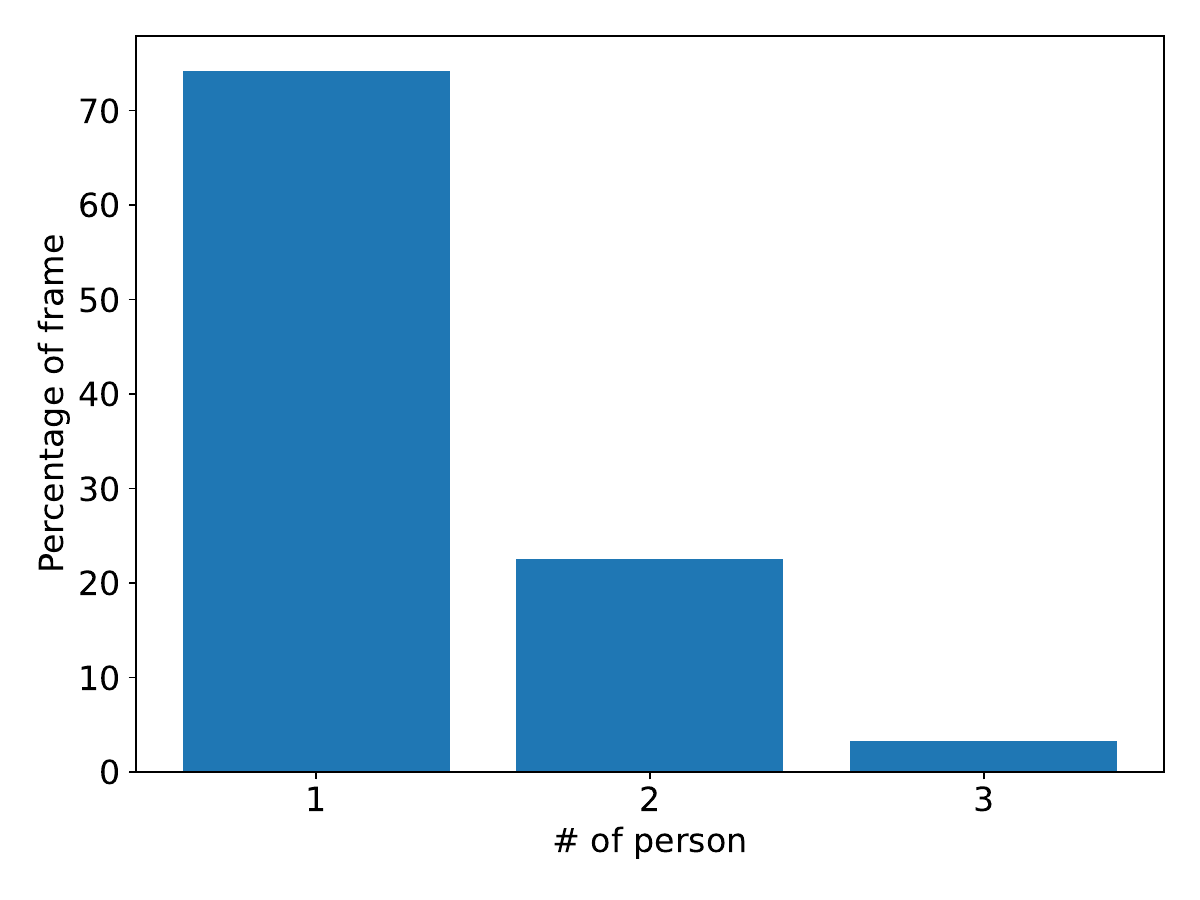}
        \caption{
            The distribution of the number of subjects over data frames.
        }
        \label{fig:num_sbj_histogram}
    \end{minipage}
\end{figure}

\begin{figure}[t]
    \centering
    \includegraphics[width=0.75\textwidth]{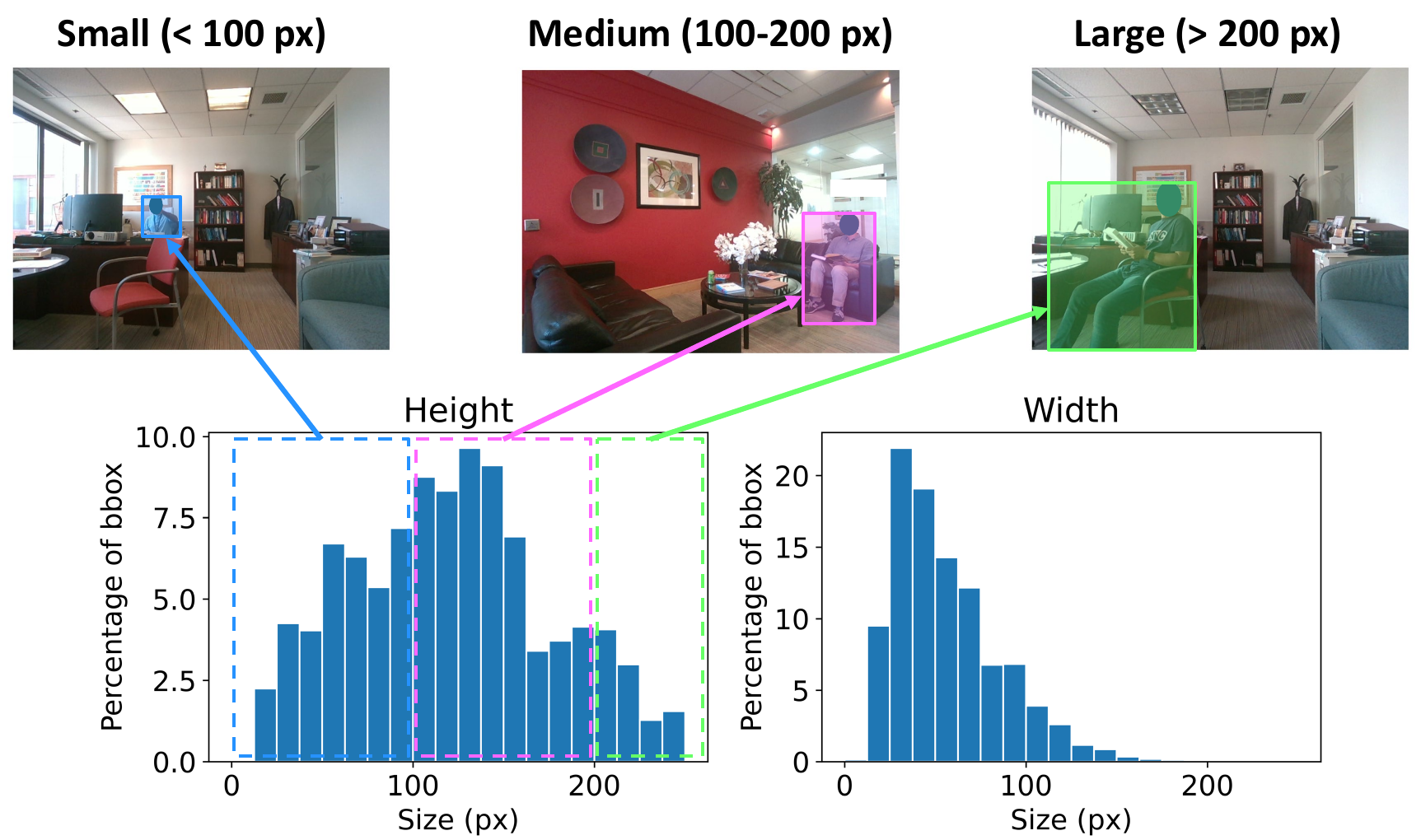}
    \caption{The distribution of annotated bounding box sizes (height and width).}
    \label{fig:hist_of_bbox_size}
\end{figure}

\begin{figure}
    \centering
    \includegraphics[width=\textwidth]{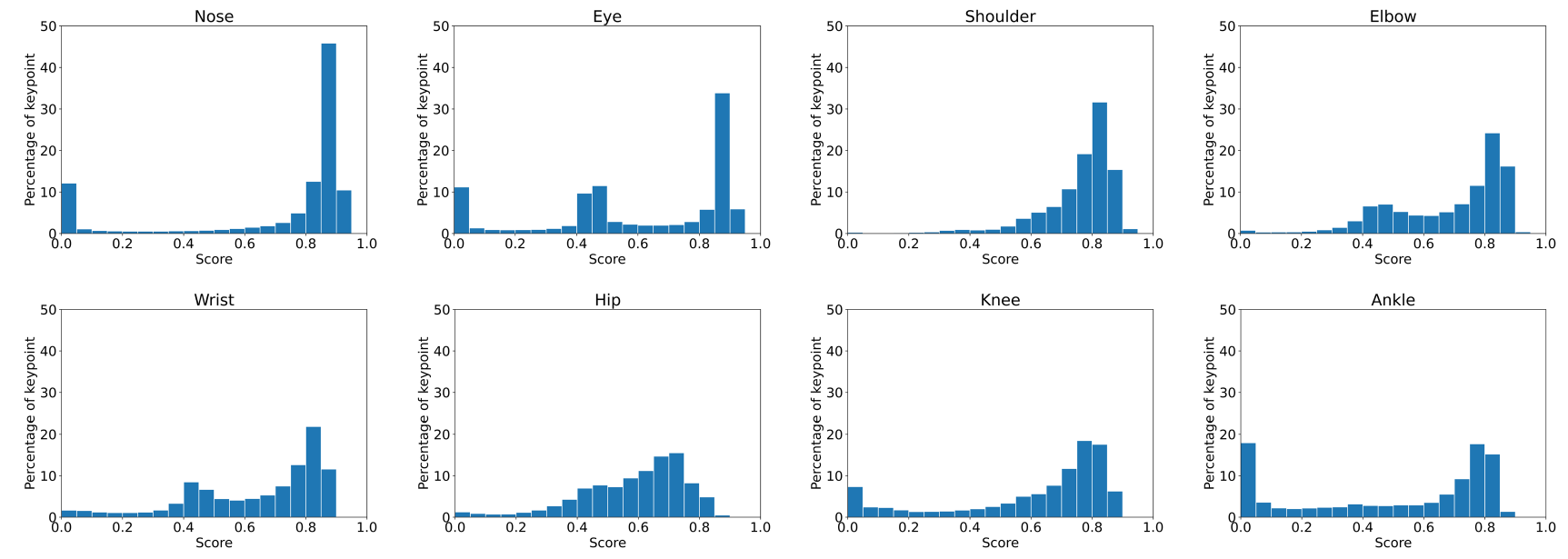}
    \caption{The distribution of keypoint annotation confidence scores over selected joints, e.g., nose, shoulder, hip, etc..}
    \label{fig:KP_score_breakdown}
\end{figure}

\section{Statistics of Annotation Labels}
\label{app:statistics}
Fig.~\ref{fig:num_annotation_per_sess} shows the average number of annotation labels for each data session between d5 and d9. Each color represents a session number (s1-s6) within a day. It is seen that earlier sessions, e.g., s1, s2, s3, are mostly single-subject sessions, while later sessions have multiple subjects. The distribution of single-person, two-person, and three-person annotations over data frames is shown in Fig.~\ref{fig:num_sbj_histogram}.

Fig.~\ref{fig:hist_of_bbox_size} further analyzes the distribution of annotated BBox sizes (height and width) in terms of pixel numbers. We categorize the BBoxes into small, medium, and large boxes in the height domain with a snapshot of each category shown in the picture above the distribution. This distribution indicates that the BBox annotations are well-balanced with a variety of distances between the camera and the subjects and sufficient diversity of subject postures.

Fig.~\ref{fig:KP_score_breakdown} breaks down the keypoint annotation confidence scores over selected joints or body parts.
Note that the displayed scores represent the average values for the left and right sides of all body parts except the nose.
With human curation involved, Fig.~\ref{fig:KP_score_breakdown} suggests that our annotation process yields high-confidence labels across the board.
However, it's noteworthy that body parts located on the torso are annotated with significantly greater confidence compared to limbs and facial parts.
Specifically, the nose, eyes, knees, and ankles exhibit a pronounced peak around zero confidence.
This lower confidence level primarily arises from these parts being prone to occlusion, such as when a person is facing away from the camera or is seated behind a desk.

\section{Synchronization}
\label{app:sync}

\begin{figure}[t]
    \centering
    \includegraphics[width=\textwidth]{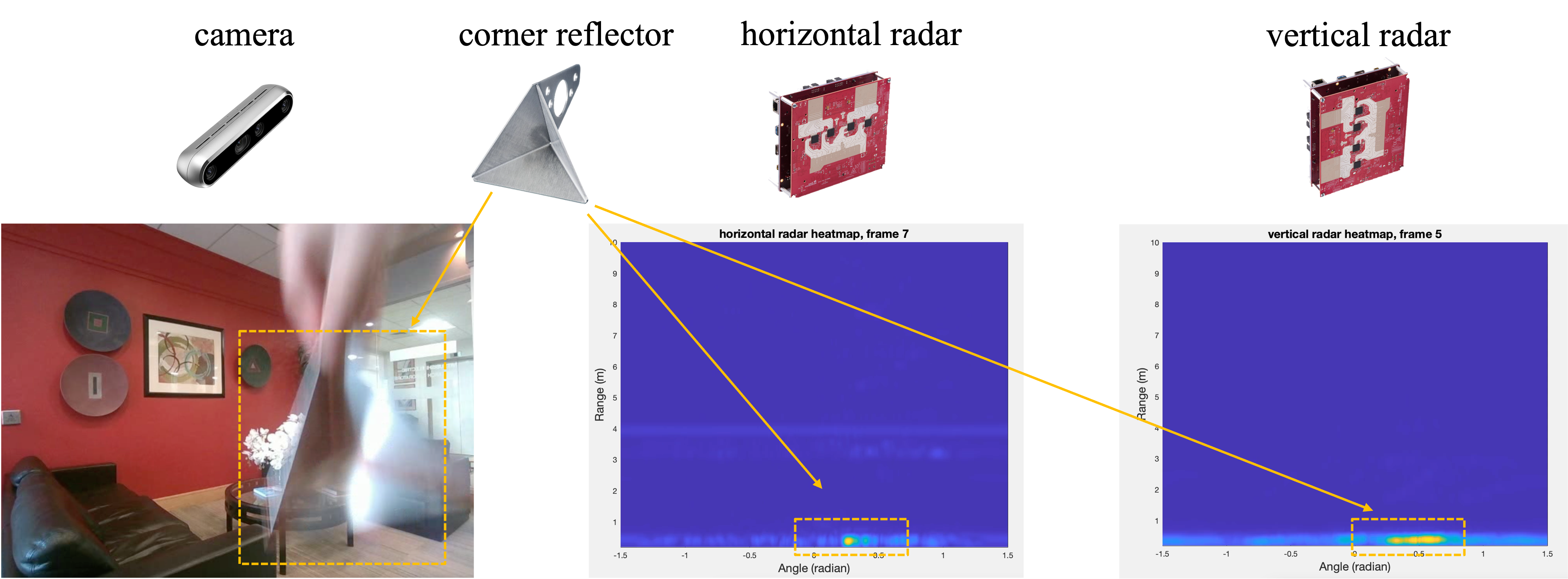} 
    \hfill 
    \caption{Initial alignment between camera and two radar sensors using a corner reflector. } 
    \label{fig:sync}
    \vspace{-0.1in}
\end{figure}

The initial alignment between the camera and two radar sensors is done with the use of a passive device, a corner reflector. A corner reflector is a passive device designed to reflect radar waves back toward the radar transceivers, regardless of the angle of incidence, with a geometric shape with three mutually perpendicular, flat surfaces or planes. As an ideal target for calibration and testing of radar sensors, we place it in front of our MMVR sensor testbed such that both camera and radar sensors can identify its presence at the beginning of each data collection session. 

One example is shown in Fig.~\ref{fig:sync} for the data session $d5s1$. The corner reflector is visualized in an RGB image frame which is identified as Frame $0$ for the image stream. Meanwhile, we preprocess the radar raw waveform, obtain the range-azimuth heatmap for the horizontal radar,  and identify the frame with strong reflection (bright yellow) at the close distance (less than $50$ cm) as Frame $0$ for the vertical radar. A similar procedure is done for the vertical radar in the range-elevation heatmap. Once identifying Frame $0$ for the camera and two radar sensors, the consequent frames are automatically aligned due to the use of the same frame rate of $15$ fps. For the refined alignment, please refer to the enclosed demo videos. 

\section{Details about Data Splits \textbf{S1} and \textbf{S2}}
\label{app:datasettings}
Table~\ref{tab:detailed_dataset_splits_for_eval} presents the detailed list of data segments within the two data splits \textbf{S1} and \textbf{S2} under both protocols \textbf{P1} and \textbf{P2}. These non-overlapping data segments were randomly selected once and then fixed to training, validation, and test sets, following the procedure outlined in Sec.~\ref{sec:setup}. We highlight that, for \textbf{S2} under the multi-room data collection \textbf{P2}, we exclude all data segments of d8 from the training and validation sets such that \textbf{S2} can be used to assess the perception performance under an unseen environment.

\begin{table}[t]
    \centering
    \caption{The detailed list of data segments for two data splits \textbf{S1} and \textbf{S2} under the two data collection protocols \textbf{P1} and \textbf{P2}. }
    \setlength\tabcolsep{4pt}
    \resizebox{0.95\columnwidth}{!}{ %
    \begin{tabular}{p{2cm}|p{2cm}| p{2cm}|p{20cm}}
        \toprule
        \textbf{Protocol} & \textbf{Split} & \textbf{Type} & \textbf{Data Sessions/Data Segments} \\
        \midrule
        \multirow{20}{*}{\textbf{P1}}  & \multirow{10}{*}{\textbf{S1}}   & \multirow{8}{*}{Training} & d1s1/000, d1s1/001, d1s1/002, d1s1/003, d1s1/004, d1s1/005, d1s1/008, d1s1/011, d1s1/012, d1s1/013, d1s1/014, d1s1/015, d1s1/016, d1s1/017, d1s1/018, d1s1/019, d1s1/020, d1s1/021, d1s1/022, d1s1/023, d1s1/024, d1s1/025, d1s1/026, d1s1/027, d1s1/028, d1s1/029, d1s1/030, d1s1/031, d1s1/032, d1s1/033, d1s1/034, d1s1/035, d1s1/036, d1s2/000, d1s2/001, d1s2/002, d1s2/004, d1s2/005, d1s2/006, d1s2/007, d1s2/008, d1s2/009, d1s2/010, d1s2/011, d1s2/012, d1s2/013, d1s2/014, d1s2/015, d1s2/017, d1s2/018, d1s2/019, d2s2/000, d2s2/001, d2s2/002, d2s2/003, d2s2/004, d2s2/005, d2s2/008, d2s2/009, d2s2/010, d2s2/011, d2s2/012, d2s2/013, d2s2/015, d2s2/016, d2s2/017, d2s2/018, d2s2/019, d3s1/000, d3s1/001, d3s1/002, d3s1/003, d3s1/004, d3s1/005, d3s1/006, d3s1/007, d3s1/010, d3s1/014, d3s1/015, d3s1/017, d3s1/018, d3s2/000, d3s2/004, d3s2/005, d3s2/006, d3s2/007, d3s2/008, d4s1/000, d4s1/001, d4s1/002, d4s1/003, d4s1/005, d4s1/007, d4s1/008, d4s1/009, d4s1/010, d4s1/011, d4s1/012. \\
        \cline{3-4}
         & & Validation & d1s1/009, d1s1/010, d1s1/037, d1s2/016, d2s2/006, d3s1/008, d3s1/012, d3s1/016, d3s2/002, d3s2/003, d4s1/004, d4s1/014 \\
         \cline{3-4}
         &  & Test & d1s1/006, d1s1/007, d1s1/038, d1s2/003, d2s2/007, d2s2/014, d3s1/009, d3s1/011, d3s1/013, d3s2/001, d4s1/006, d4s1/013 \\
         \cline{2-4}
        & \multirow{10}{*}{\textbf{S2}} & \multirow{8}{*}{Training} & d1s1/000, d1s1/001, d1s1/002, d1s1/003, d1s1/004, d1s1/005, d1s1/006, d1s1/007, d1s1/008, d1s1/009, d1s1/010, d1s1/011, d1s1/012, d1s1/013, d1s1/014, d1s1/015, d1s1/016, d1s1/017, d1s1/018, d1s1/019, d1s1/020, d1s1/021, d1s1/022, d1s1/023, d1s1/024, d1s1/025, d1s1/026, d1s1/027, d1s1/028, d1s1/029, d1s1/030, d1s1/031, d1s1/032, d1s1/033, d1s1/034, d1s1/035, d1s1/036, d1s1/037, d1s1/038, d1s2/000, d1s2/001, d1s2/002, d1s2/003, d1s2/004, d1s2/005, d1s2/006, d1s2/007, d1s2/008, d1s2/009, d1s2/010, d1s2/011, d1s2/012, d1s2/013, d1s2/014, d1s2/015, d1s2/016, d1s2/017, d1s2/018, d1s2/019, d2s2/000, d2s2/001, d2s2/002, d2s2/003, d2s2/004, d2s2/005, d2s2/006, d2s2/007, d2s2/008, d2s2/009, d2s2/010, d2s2/011, d2s2/012, d2s2/013, d2s2/014, d2s2/015, d2s2/016, d2s2/017, d2s2/018, d2s2/019 \\
         \cline{3-4}
         &  & \multirow{3}{*}{Validation} & d3s1/000, d3s1/001, d3s1/002, d3s1/003, d3s1/004, d3s1/005, d3s1/006, d3s1/007, d3s1/008, d3s1/009, d3s1/010, d3s1/011, d3s1/012, d3s1/013, d3s1/014, d3s1/015, d3s1/016, d3s1/017, d3s1/018, d3s2/000, d3s2/001, d3s2/002, d3s2/003, d3s2/004, d3s2/005, d3s2/006, d3s2/007, d3s2/008 \\
         \cline{3-4}
         &  & \multirow{2}{*}{Test} & d4s1/000, d4s1/001, d4s1/002, d4s1/003, d4s1/004, d4s1/005, d4s1/006, d4s1/007, d4s1/008, d4s1/009, d4s1/010, d4s1/011, d4s1/012, d4s1/013, d4s1/014 \\
         \midrule
         \multirow{45}{*}{\textbf{P2}} & \multirow{25}{*}{\textbf{S1}} & \multirow{18}{*}{Training} & d5s1/001, d5s1/002, d5s1/003, d5s1/004, d5s1/005, d5s1/007, d5s1/008, d5s2/000, d5s2/001, d5s2/002, d5s2/005, d5s2/006, d5s2/007, d5s2/008, d5s3/001, d5s3/003, d5s3/004, d5s3/005, d5s3/006, d5s3/007, d5s3/008, d5s3/009, d5s4/000, d5s4/001, d5s4/003, d5s4/004, d5s4/005, d5s4/006, d5s5/000, d5s5/002, d5s5/004, d5s6/000, d5s6/001, d5s6/002, d5s6/003, d5s6/004, d5s6/005, d5s6/007, d5s6/008, d5s6/009, d6s1/000, d6s1/001, d6s1/002, d6s1/003, d6s1/004, d6s1/005, d6s1/006, d6s1/007, d6s1/008, d6s1/009, d6s2/000, d6s2/001, d6s2/003, d6s2/005, d6s2/008, d6s2/009, d6s3/000, d6s3/001, d6s3/002, d6s3/003, d6s3/005, d6s3/007, d6s3/009, d6s4/000, d6s4/001, d6s4/002, d6s4/003, d6s4/004, d6s4/005, d6s4/006, d6s4/007, d6s4/008, d6s5/000, d6s6/000, d6s6/001, d6s6/002, d6s6/003, d6s6/004, d6s6/005, d6s6/006, d6s6/009, d7s1/000, d7s1/001, d7s1/003, d7s1/004, d7s1/005, d7s1/006, d7s1/007, d7s1/009, d7s2/000, d7s2/001, d7s2/003, d7s2/004, d7s2/005, d7s2/006, d7s2/007, d7s2/008, d7s2/009, d7s3/000, d7s3/001, d7s3/003, d7s3/005, d7s3/006, d7s3/007, d7s3/008, d7s3/009, d7s4/000, d7s4/001, d7s4/002, d7s4/003, d7s4/004, d7s4/005, d7s4/006, d7s4/007, d7s4/008, d7s5/001, d7s5/002, d7s5/003, d7s5/004, d7s5/005, d7s5/007, d7s5/008, d7s5/009, d8s1/000, d8s1/001, d8s1/002, d8s1/003, d8s1/004, d8s1/005, d8s1/006, d8s1/007, d8s1/008, d8s1/009, d8s2/000, d8s2/002, d8s2/003, d8s2/004, d8s2/005, d8s2/006, d8s2/007, d8s2/008, d8s2/009, d8s3/000, d8s3/001, d8s3/002, d8s3/003, d8s3/004, d8s3/006, d8s3/007, d8s3/008, d8s3/009, d8s4/001, d8s4/002, d8s4/003, d8s4/004, d8s4/005, d8s4/006, d8s4/007, d8s4/008, d8s4/009, d8s5/000, d8s5/001, d8s5/003, d8s5/004, d8s5/005, d8s5/006, d8s5/008, d8s6/000, d8s6/001, d8s6/003, d8s6/004, d8s6/005, d8s6/006, d8s6/007, d8s6/009, d9s1/001, d9s1/002, d9s1/005, d9s1/006, d9s1/007, d9s1/008, d9s2/000, d9s2/001, d9s2/003, d9s2/004, d9s2/005, d9s2/006, d9s2/007, d9s2/008, d9s3/001, d9s3/003, d9s3/004, d9s3/005, d9s3/006, d9s3/007, d9s3/008, d9s3/009, d9s4/001, d9s4/002, d9s4/003, d9s4/006, d9s4/007, d9s5/000, d9s5/001, d9s5/002, d9s5/003, d9s5/004, d9s5/005, d9s5/006, d9s5/007, d9s5/008, d9s5/009, d9s6/000, d9s6/001, d9s6/002, d9s6/003, d9s6/006, d9s6/007, d9s6/008 \\
        \cline{3-4}
         &  &  \multirow{3}{*}{Validation} & d5s1/000, d5s1/006, d5s2/003, d5s2/004, d5s3/000, d5s4/002, d5s4/008, d5s5/001, d5s5/005, d6s2/002, d6s3/004, d6s3/008, d6s6/007, d6s6/008, d7s2/002, d7s5/000, d8s3/005, d8s5/002, d9s1/004, d9s2/002, d9s3/000, d9s3/002, d9s4/000, d9s4/004, d9s4/008, d9s4/009, d9s6/005 \\
         \cline{3-4}
         &  &  \multirow{3}{*}{Test} & d5s3/002, d5s4/007, d5s5/003, d5s6/006, d6s2/004, d6s2/006, d6s2/007, d6s3/006, d6s4/009, d7s1/002, d7s1/008, d7s3/002, d7s3/004, d7s4/009, d7s5/006, d8s2/001, d8s4/000, d8s5/007, d8s5/009, d8s6/002, d8s6/008, d9s1/000, d9s1/003, d9s2/009, d9s4/005, d9s6/004, d9s6/009 \\
         \cline{2-4}
         & \multirow{22}{*}{\textbf{S2}} &  \multirow{12}{*}{Training} & d5s3/000, d5s3/001, d5s3/002, d5s3/003, d5s3/004, d5s3/005, d5s3/006, d5s3/007, d5s3/008, d5s3/009, d5s4/000, d5s4/001, d5s4/002, d5s4/003, d5s4/004, d5s4/005, d5s4/006, d5s4/007, d5s4/008, d5s5/000, d5s5/001, d5s5/002, d5s5/003, d5s5/004, d5s5/005, d5s6/000, d5s6/001, d5s6/002, d5s6/003, d5s6/004, d5s6/005, d5s6/006, d5s6/007, d5s6/008, d5s6/009, d6s2/000, d6s2/001, d6s2/002, d6s2/003, d6s2/004, d6s2/005, d6s2/006, d6s2/007, d6s2/008, d6s2/009, d6s3/000, d6s3/001, d6s3/002, d6s3/003, d6s3/004, d6s3/005, d6s3/006, d6s3/007, d6s3/008, d6s3/009, d6s5/000, d6s6/000, d6s6/001, d6s6/002, d6s6/003, d6s6/004, d6s6/005, d6s6/006, d6s6/007, d6s6/008, d6s6/009, d7s1/000, d7s1/001, d7s1/002, d7s1/003, d7s1/004, d7s1/005, d7s1/006, d7s1/007, d7s1/008, d7s1/009, d7s3/000, d7s3/001, d7s3/002, d7s3/003, d7s3/004, d7s3/005, d7s3/006, d7s3/007, d7s3/008, d7s3/009, d7s4/000, d7s4/001, d7s4/002, d7s4/003, d7s4/004, d7s4/005, d7s4/006, d7s4/007, d7s4/008, d7s4/009, d9s1/000, d9s1/001, d9s1/002, d9s1/003, d9s1/004, d9s1/005, d9s1/006, d9s1/007, d9s1/008, d9s2/000, d9s2/001, d9s2/002, d9s2/003, d9s2/004, d9s2/005, d9s2/006, d9s2/007, d9s2/008, d9s2/009, d9s4/000, d9s4/001, d9s4/002, d9s4/003, d9s4/004, d9s4/005, d9s4/006, d9s4/007, d9s4/008, d9s4/009, d9s5/000, d9s5/001, d9s5/002, d9s5/003, d9s5/004, d9s5/005, d9s5/006, d9s5/007, d9s5/008, d9s5/009 \\
         \cline{3-4}
         &  &  \multirow{3}{*}{Validation} & d5s2/000, d5s2/001, d5s2/002, d5s2/003, d5s2/004, d5s2/005, d5s2/006, d5s2/007, d5s2/008, d6s4/000, d6s4/001, d6s4/002, d6s4/003, d6s4/004, d6s4/005, d6s4/006, d6s4/007, d6s4/008, d6s4/009, d7s5/000, d7s5/001, d7s5/002, d7s5/003, d7s5/004, d7s5/005, d7s5/006, d7s5/007, d7s5/008, d7s5/009, d9s6/000, d9s6/001, d9s6/002, d9s6/003, d9s6/004, d9s6/005, d9s6/006, d9s6/007, d9s6/008, d9s6/009 \\
         \cline{3-4}
         &  & \multirow{8}{*}{Test} & d5s1/000, d5s1/001, d5s1/002, d5s1/003, d5s1/004, d5s1/005, d5s1/006, d5s1/007, d5s1/008, d6s1/000, d6s1/001, d6s1/002, d6s1/003, d6s1/004, d6s1/005, d6s1/006, d6s1/007, d6s1/008, d6s1/009, d7s2/000, d7s2/001, d7s2/002, d7s2/003, d7s2/004, d7s2/005, d7s2/006, d7s2/007, d7s2/008, d7s2/009, d9s3/000, d9s3/001, d9s3/002, d9s3/003, d9s3/004, d9s3/005, d9s3/006, d9s3/007, d9s3/008, d9s3/009, d8s1/000, d8s1/001, d8s1/002, d8s1/003, d8s1/004, d8s1/005, d8s1/006, d8s1/007, d8s1/008, d8s1/009, d8s2/000, d8s2/001, d8s2/002, d8s2/003, d8s2/004, d8s2/005, d8s2/006, d8s2/007, d8s2/008, d8s2/009, d8s3/000, d8s3/001, d8s3/002, d8s3/003, d8s3/004, d8s3/005, d8s3/006, d8s3/007, d8s3/008, d8s3/009, d8s4/000, d8s4/001, d8s4/002, d8s4/003, d8s4/004, d8s4/005, d8s4/006, d8s4/007, d8s4/008, d8s4/009, d8s5/000, d8s5/001, d8s5/002, d8s5/003, d8s5/004, d8s5/005, d8s5/006, d8s5/007, d8s5/008, d8s5/009, d8s6/000, d8s6/001, d8s6/002, d8s6/003, d8s6/004, d8s6/005, d8s6/006, d8s6/007, d8s6/008, d8s6/009 \\
        \bottomrule
    \end{tabular}
    \label{tab:detailed_dataset_splits_for_eval}
    }
\end{table}

\section{Details of Evaluation Hyper-Parameters and Baseline Methods}
\label{app:methods_prameters}

\subsection{Evaluation Hyper-Parameters}
Table~\ref{tab:detailed_hyperparameters} lists detailed information on the data and training hyper-parameters.
For the data part, we list the number of data frames for training, validation, and test sets under each evaluation scenario, the input radar heatmap size, segmentation mask size, and the output size used to compute the training loss. For the training part, we include hyper-parameters such as batch size, number of epochs, parameters related to early stopping, learning rate, etc. 
\begin{table}[t]
    \centering
    \footnotesize
    \caption{Details about evaluation parameters.}
    \begin{tabular}{c|c|c|lcc}
        \toprule
         & \multicolumn{3}{c}{\textbf{Name}} & \textbf{Notation} & \textbf{Value} \\
        \midrule
        \multirow{19}{*}{\rotatebox{90}{\textbf{Data}}} & \multirow{6}{*}{\textbf{P1}} & \multirow{3}{*}{\textbf{S1}} & \# of training & - & 86579 \\
         &  &  & \# of validation & - & 10538 \\
         &  &  & \# of test & - & 10785 \\
         \cline{3-6}
         &  & \multirow{3}{*}{\textbf{S2}} & \# of training & - & 70266 \\
         &  &  & \# of validation & - & 24398 \\
         &  &  & \# of test & - & 13238 \\
         \cline{2-6}
         & \multirow{6}{*}{\textbf{P2}} & \multirow{3}{*}{\textbf{S1}} & \# of training & - & 190441 \\
         &  &  & \# of validation & - & 23899 \\
         &  &  & \# of test & - & 23458 \\
         \cline{3-6}
         &  & \multirow{3}{*}{\textbf{S2}} & \# of training & - & 118280 \\
         &  &  & \# of validation & - & 33841 \\
         &  &  & \# of test & - & 85677 \\
         \cline{2-6}
         % & \multicolumn{3}{l}{Original radar heatmap size} & $H\times W$ & 512$\times$256 \\
         & \multicolumn{3}{l}{Input radar heatmap size} & $H\times W$ & 256$\times$128 \\
         & \multicolumn{3}{l}{Original segmentation mask size} & $H\times W$ & 480$\times$640 \\
         & \multicolumn{3}{l}{Input segmentation mask size} & $H\times W$ & 240$\times$320 \\
         & \multicolumn{3}{l}{Fixed-height size (pixel)} & $H$ & 36 \\
         & \multicolumn{3}{l}{Heatmap/PAF size for pose estimation} & $H\times W$ & 46$\times$46 \\
         & \multicolumn{3}{l}{Scale} & - & log \\
        \midrule
        \multirow{8}{*}{\rotatebox{90}{\textbf{Training}}} & \multicolumn{3}{l}{batch size} & - & 32 \\
         & \multicolumn{3}{l}{epoch} & - & 100 \\
         & early stopping & \multicolumn{2}{l}{patience} & - & 5 \\
         & early stopping & \multicolumn{2}{l}{check val every $N$ epoch} & - & 2 \\
         & \multicolumn{3}{l}{optimizer} & - & Adam \\
         & \multicolumn{3}{l}{learning rate} & - & 1e-4 \\
         & \multicolumn{3}{l}{sheduler Linear $\times$ 0.1} & - & 50 \\
         & \multicolumn{3}{l}{weight decay} & - & 1e-4 \\
        \bottomrule
    \end{tabular}
    \label{tab:detailed_hyperparameters}
\end{table}

\begin{figure}[t]
    \begin{tabular}{c}
        \begin{minipage}[t]{\hsize}
            \centering
            \includegraphics[width=0.75\textwidth]{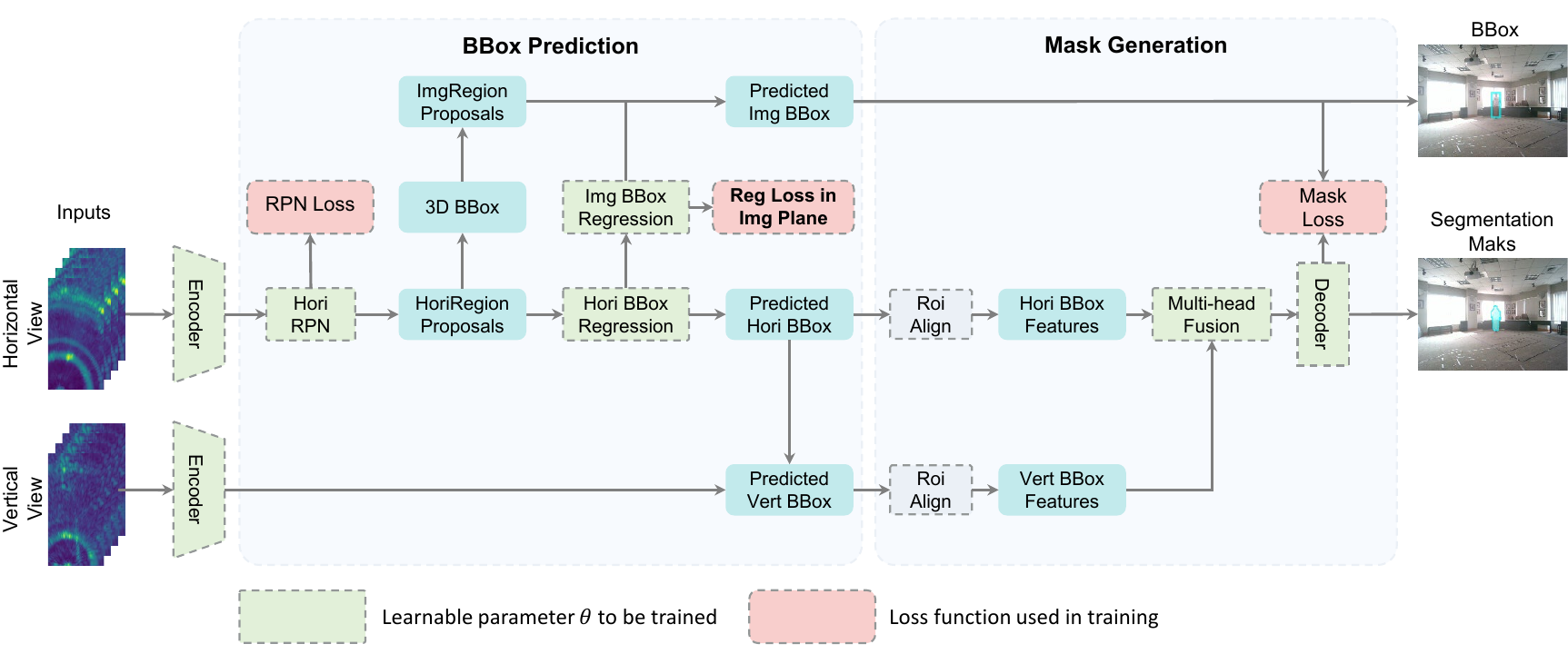}
            \caption{Baseline method for object detection and segmentation: RFMask~(\cite{RFMask23}). }
            \label{fig:arch_rfmask}
        \end{minipage} \\
        \begin{minipage}[t]{\hsize}
            \centering
            \includegraphics[width=0.55\textwidth]{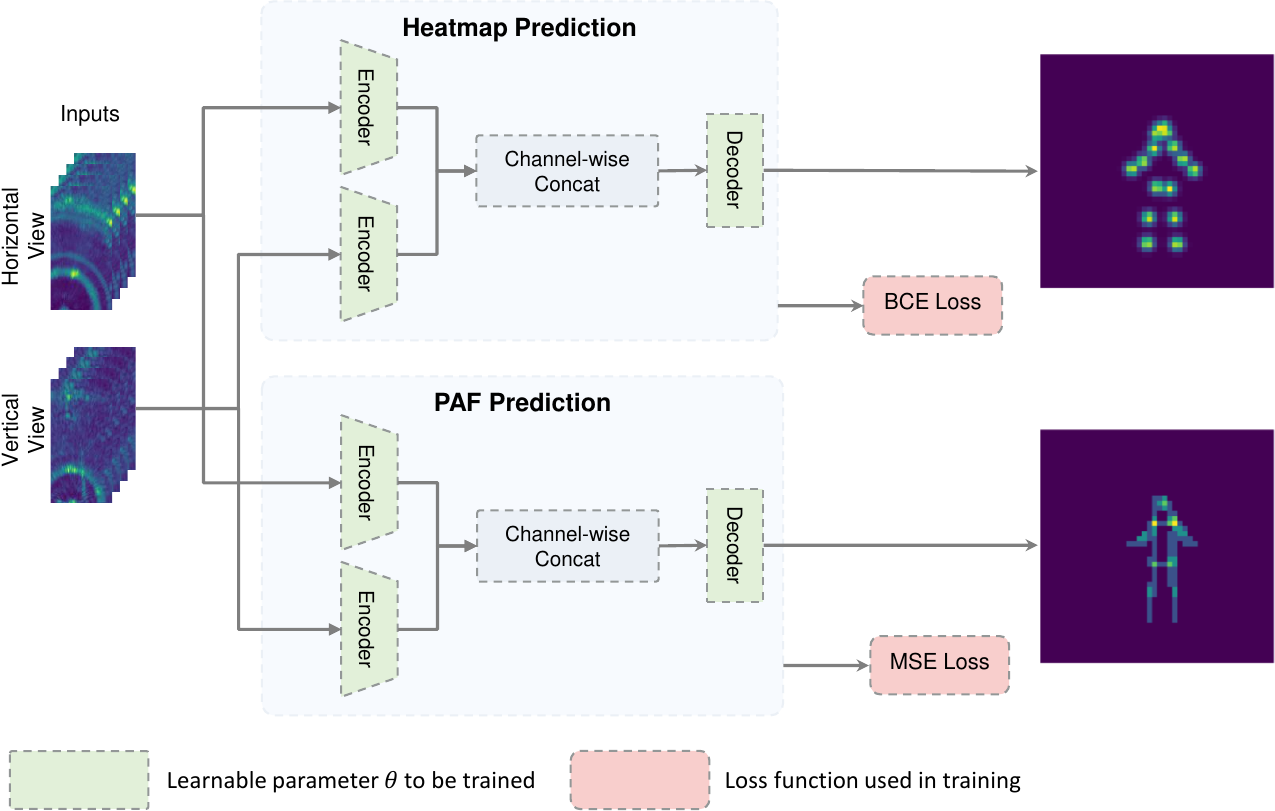}
            \caption{Baseline method for pose estimation: RF-Pose~(\cite{RFPose18}). }
            \label{fig:arch_rfpose}
        \end{minipage} \\
        \begin{minipage}[t]{\hsize}
            \centering
            \includegraphics[width=0.75\textwidth]{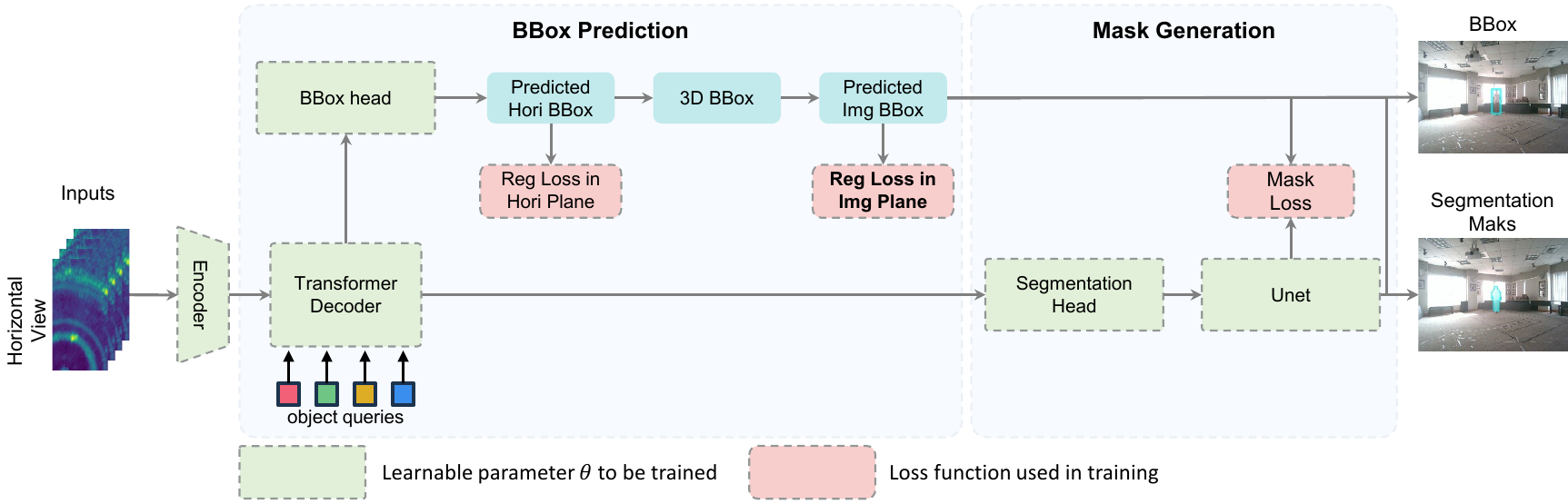}
            \caption{Additional baseline for Object Detection and segmentation: DETR~(\cite{Carion2020_detr}). }
            \label{fig:arch_detr}
        \end{minipage}
    \end{tabular}
\end{figure}

\subsection{Baseline Methods for Object Detection and Segmentation}
We use RFMask~(\cite{RFMask23}) as one baseline method for BBox and segmentation tasks. Unlike the HIBER dataset with the BBox labels on the two radar views, our dataset MMVR annotates the BBox labels directly on the image plane. As a result, we modify RFMask in a way that the BBox loss is calculated directly on the image plane and backpropagates to trainable parameters in an end-to-end fashion. 

Fig.~\ref{fig:arch_rfmask} shows the modified RFMask architecture with the loss function computed directly on the image plane. Specifically, we add an \texttt{Img BBox Regression} module alongside a \texttt{Hori BBox Regression} module, enabling the conversion of BBox offsets to the image plane. By computing loss with respect to these offsets, we can directly learn BBoxes on the image plane. Additionally, the region proposals estimated by the region proposal network (RPN) are transformed into 3D BBoxes based on the fixed-height size (see Table~\ref{tab:detailed_hyperparameters}), the same as the original RFMask. These 3D BBoxes are then projected onto the image plane using the Camera-Radar calibration of Sec.~\ref{calibration} and a 3D-to-2D projection using  intrinsic parameters obtained from the RealSense camera D455. 

In Sec.~\ref{sec:ablation}, we also consider an additional baseline method that borrows the more advanced Detection Transformer (DETR)~(\cite{Carion2020_detr}) for the radar perception tasks of object detection and instance segmentation. The DETR-based architecture is shown in Fig.~\ref{fig:arch_detr}, with a modification to convert BBoxes predicted on the horizontal plane to the image plane. The BBoxe loss is calculated on both the horizontal and image planes. For the segmentation task, UNet is added and the mask loss is computed. During the training, only the parameters of the Transformer Decoder and BBox head are initially updated, and then, the parameters of the segmentation head and UNet are updated while keeping the parameters of the Transformer Decoder and BBox head fixed.

\subsection{Baseline Method for Pose Estimation}
We use RF-Pose~(\cite{RFPose18}) as a baseline method for the evaluation of pose estimation performance on our dataset MMVR. The original RF-Pose aims at predicting only a confidence heatmap for each joint, which is then used for multi-person association. We extend its architecture by using two encoder-decoder sets to predict both heatmap and part affinity map (PAF), to follow the common practice in the bottom-up pose estimation in computer vision and provide better multi-person association~(\cite{OpenPose17}). Fig.~\ref{fig:arch_rfpose} shows our implementation of RF-Pose with the exact the same encoder and decoder structures as RF-Pose~(\cite[Sec.~4.3]{RFPose18}) (e.g., 3D convolution, stride and convolution kernel parameters) with the loss function combining both the keypoint heatmap and PAF. 

\section{Additional Visualization Results}
\label{app:visualization}
We provide additional qualitative inspection of results in 
Fig.~\ref{fig:vis_detection} for object detection, Fig.~\ref{fig:vis_pose} for pose estimation, and Fig.~\ref{fig:vis_segmentation} for instance segmentation, respectively. For each perception task, we select one frame from d1 in \textbf{P1} and one frame from each of the $5$ days (d5-d9) in \textbf{P2}. Meanwhile, we show failure cases for each perception task in Fig.~\ref{fig:vis_bad_cases}.

\subsection{Object Detection}
\begin{figure}[t]
    \centering
    \includegraphics[width=\textwidth]{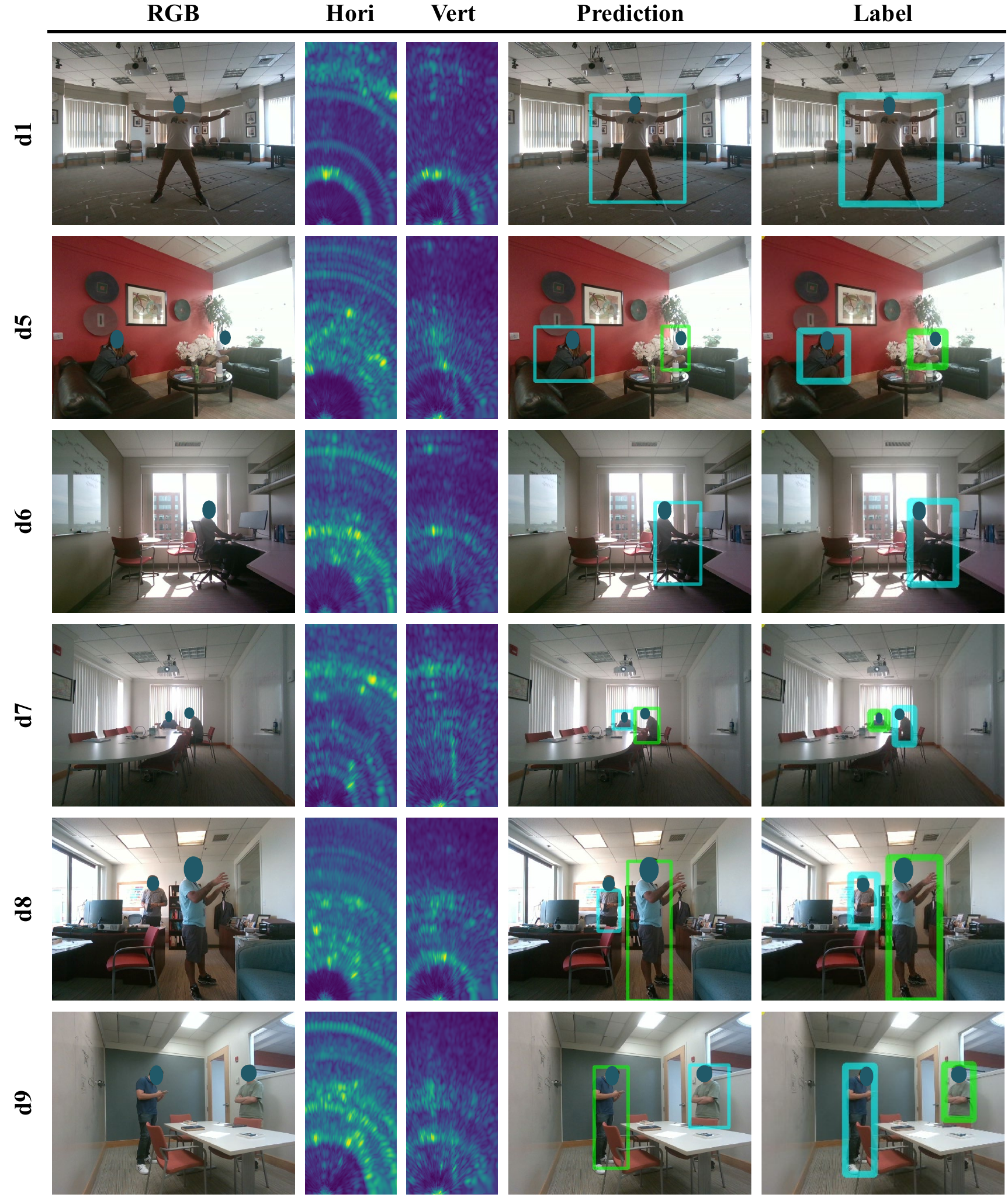}
    \caption{
        Visualization of baseline object detection results. 
    }
    \label{fig:vis_detection}
\end{figure}

As shown in Fig.~\ref{fig:vis_detection}, it is possible to localize the subject in the image plane using two radar views. The results for d1 (the first row) show that the radar features can support the BBox prediction that includes the spread arms and legs of a subject. On the other hand, the first two rows of Fig.~\ref{fig:vis_bad_cases} show missing detection when the subject is close to the radar and camera, and when the subject is stationary for a long duration of time. 

\subsection{Pose Estimation}
\begin{figure}[t]
    \centering
    \includegraphics[width=\textwidth]{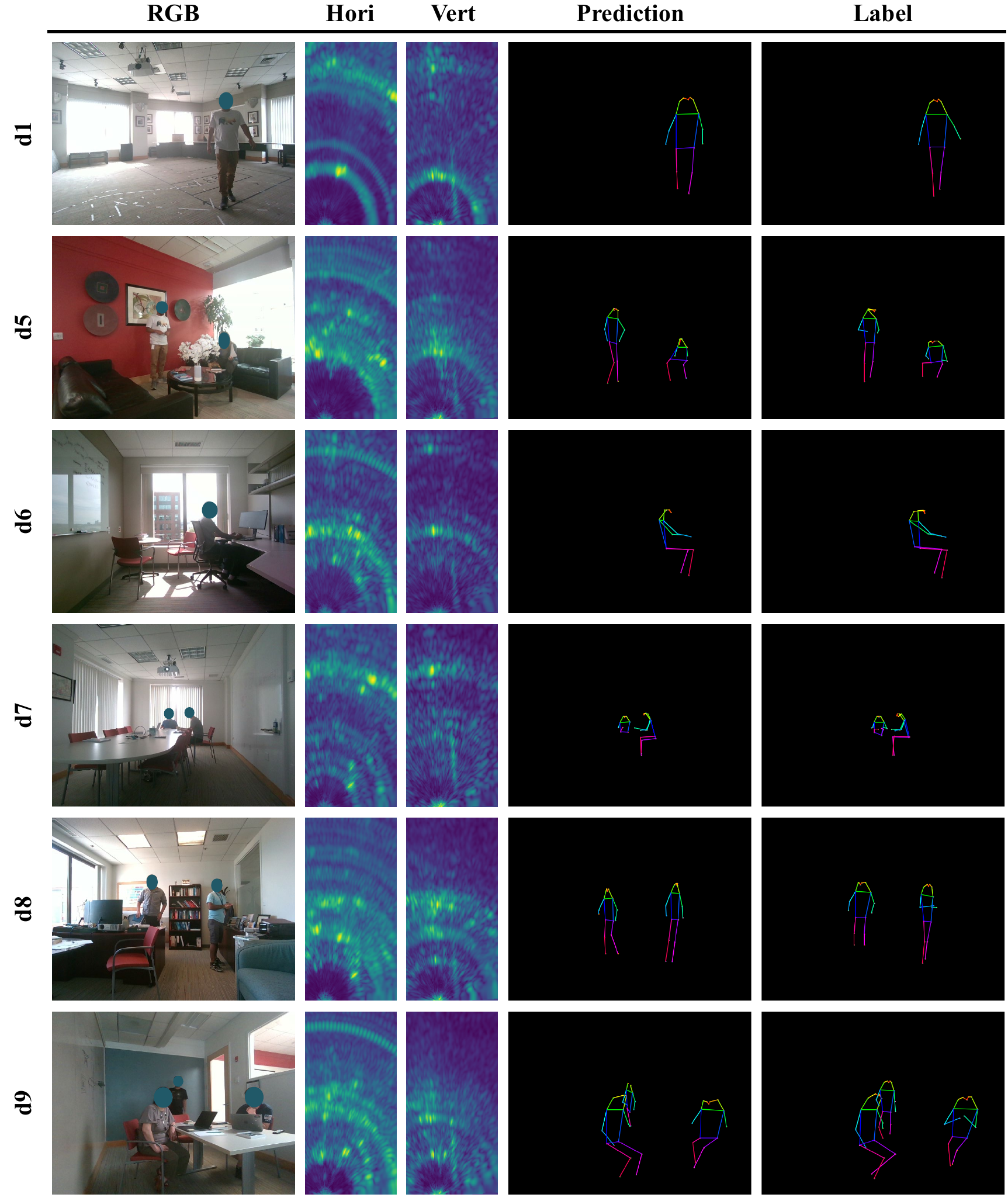}
    \caption{
        Visualization of baseline pose estimation results. 
    }
    \label{fig:vis_pose}
\end{figure}

Fig.~\ref{fig:vis_pose} shows reasonably good performance of baseline pose estimation in the single-person session such as d1 and the performance degrades in multi-person sessions. While the location and relative scale of subjects are estimated with small errors, the end of limbs like wrists and ankles tends to have larger errors, especially in more complex environments in \textbf{P2}. For the failure cases shown in the third and fourth rows of Fig.~\ref{fig:vis_bad_cases}, one can observe that certain connections between keypoints can be twisted or a subset of keypoints were missed or estimated with small confidence scores.

\subsection{Instance Segmentation}
\begin{figure}[t]
    \centering
    \includegraphics[width=\textwidth]{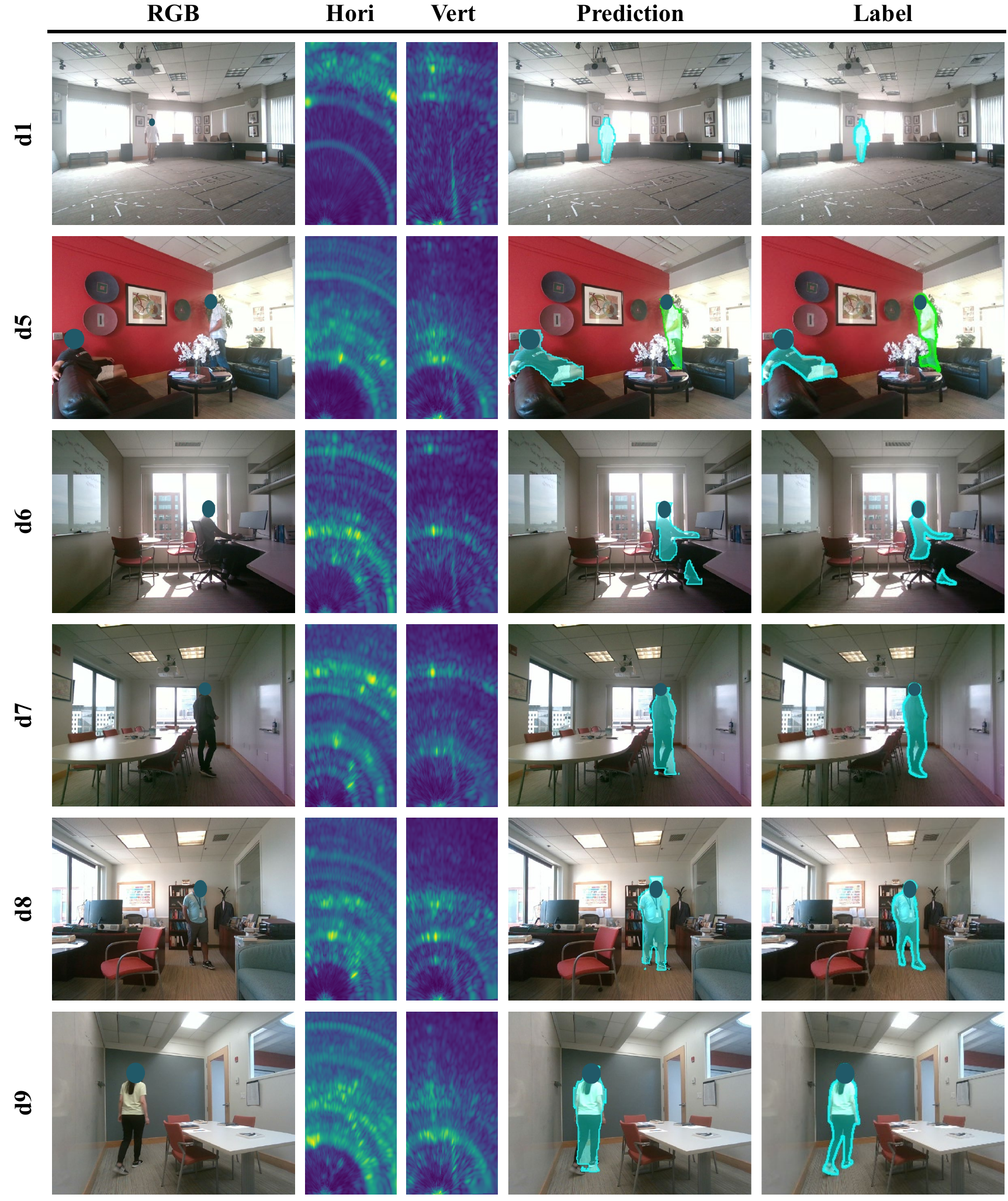}
    \caption{
        Visualization of baseline instance segmentation results. 
    }
    \label{fig:vis_segmentation}
\end{figure}

\begin{figure}[t]
    \centering
    \includegraphics[width=\textwidth]{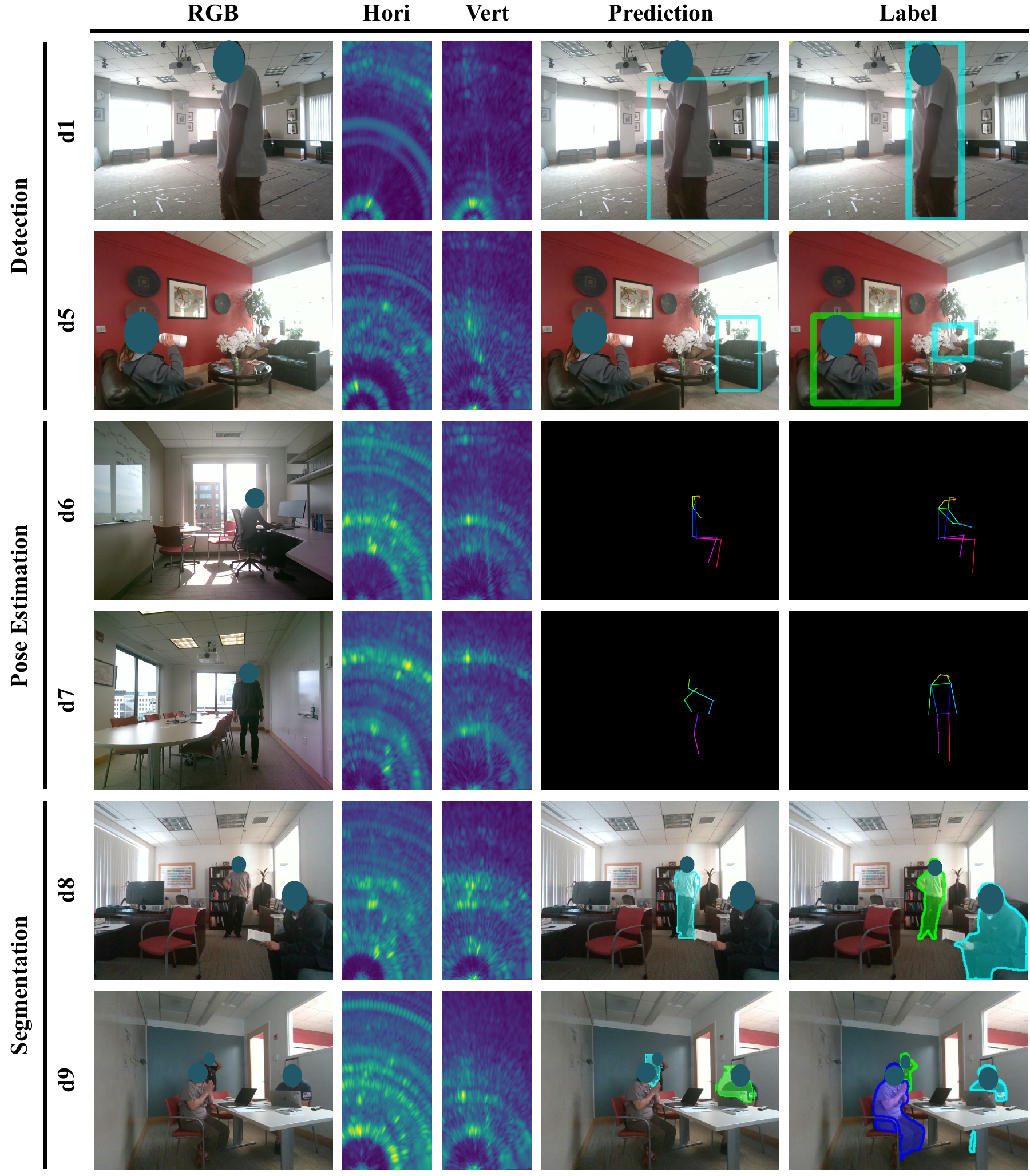}
    \caption{
        Visualization of failure cases. 
    }
    \label{fig:vis_bad_cases}
\end{figure}

Finally, for instance, segmentation results in Fig.~\ref{fig:vis_segmentation}, the baseline RFMask can capture the rough shape of subjects. On the other hand, as shown in the last two rows of Fig.~\ref{fig:vis_bad_cases}, segmentation mask predictions may be missing in the multi-person scenarios, and mask pixels in between legs can be identified as part of the predicted instance mask.  

\end{document}